%% file: main.tex
\documentclass{article}

\usepackage[preprint]{neurips_2026}
\usepackage[table]{xcolor}
\usepackage{times}
\usepackage{latexsym}
\usepackage{graphicx} 
\graphicspath{{Figure/}}
\usepackage{amsfonts}
\usepackage{amsmath}   
\usepackage{amssymb}
\usepackage{enumitem} 
\usepackage{bm}       
\usepackage{subcaption}
\usepackage{booktabs}
\usepackage{multirow}
\usepackage{tabularx}
\usepackage{makecell}
\usepackage{xspace} 
\usepackage{algorithm}
\usepackage{algorithmic}
\usepackage{booktabs}
\usepackage{pifont}
\input{preamble}
\input{macros}

\title{TRE: Training-Free Hallucination Detection for Diffusion Language Models}

\author{\textbf{Pengcheng Weng$^{1,\ast}$, Yanyu Qian$^{2,\ast}$, Yue Tan$^{3,}$\thanks{Equal Contribution.}, Yixin Liu$^{3,}$\thanks{Corresponding Author.} }\\
$^{1}$University of Bern,
$^{2}$Nanyang Technological University,
$^{3}$Griffith University\\
\texttt{pengcheng.weng@students.unibe.ch},
\texttt{yanyu003@e.ntu.edu.sg},  \\
\texttt{\{yue.tan, yixin.liu\}@griffith.edu.au}}

\begin{document}

\maketitle

\input{sections/abstract}
\input{sections/introduction}
\input{sections/related_work}

\input{sections/preliminaries}
\input{sections/method}
\input{sections/method33}
\input{sections/experiments}
\input{sections/conclusion}

{\small
\bibliographystyle{unsrt}
\bibliography{reference}
}

\clearpage
\appendix
\input{appendix/00_content}

\input{sections/Limitations_and_Broader_Impacts}
\end{document}

%% file: preamble.tex
\usepackage{amsmath,amssymb,amsfonts,amsthm,mathtools,bm}

\usepackage{graphicx}
\usepackage{subcaption}
\usepackage{booktabs}
\usepackage{multirow}
\usepackage{makecell}
\usepackage{array}
\usepackage{tabularx}
\usepackage{wrapfig}

\usepackage{enumitem}
\usepackage{xspace}
\usepackage{xcolor}
\usepackage{microtype}
\usepackage{nicefrac}
\usepackage{url}

\usepackage{hyperref}

\usepackage[nameinlink,capitalize,noabbrev]{cleveref}

\graphicspath{{figures/}}

\theoremstyle{definition}

\theoremstyle{remark}

\setlist[itemize]{leftmargin=1.2em,topsep=2pt,itemsep=1pt,parsep=0pt}
\setlist[enumerate]{leftmargin=1.4em,topsep=2pt,itemsep=1pt,parsep=0pt}

%% file: macros.tex
\newcommand{\ourmethod}{\textsc{TRE}\xspace}

\newcommand{\llada}{LLaDA\xspace}
\newcommand{\dream}{Dream\xspace}



%% file: sections/abstract.tex
\begin{abstract}
Diffusion large language models (D-LLMs) have recently gained increasing attention, yet their reliability is significantly hindered by the hallucination problem. Existing hallucination detection approaches for D-LLMs mainly follow a training-based paradigm, relying on data-driven training to optimize the detector. Such reliance not only limits their generalizability across domains models but also incurs additional training cost and deployment overhead. 
To address these limitations, we propose \ourmethod, a training-free hallucination detection metric for D-LLMs. \ourmethod is a parameter-free and single-run metric that estimates hallucination risk directly from the entropy signals of a single generation, without requiring any detector training or repeated sampling. \ourmethod extracts entropy signals within the D-LLM decoding process along both the spatial and temporal dimensions. 
From a token-level spatial perspective, we focus on revealing tokens as the most informative carriers of uncertainty, capturing where uncertainty is actively committed. From a diffusion step-level temporal perspective, we empirically identify the dominance of late-step entropy and hence aggregate these signals with a simple linear weighting scheme to obtain TRE.
Extensive experiments on multiple D-LLMs and QA datasets demonstrate that \ourmethod achieves competitive performance, while enjoying strong generalizability, efficiency, and robustness.

\end{abstract}

%% file: sections/introduction.tex
\section{Introduction}
\label{sec:intro}

\begin{figure*}[t]
\vspace{-2mm}
    \centering
    \includegraphics[width=\textwidth]{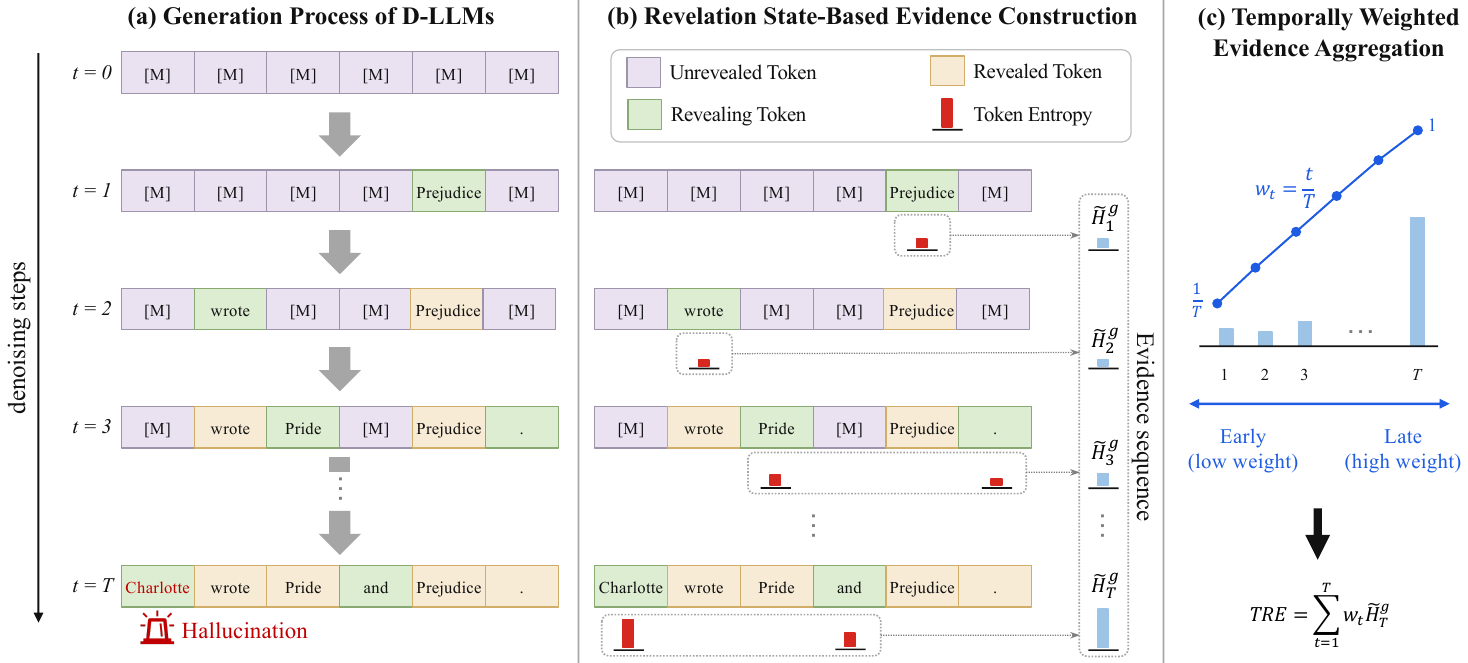}
    \vspace{-2mm}
    \caption{
    Overview of (a) the generation process of D-LLMs and the two key components of \ourmethod, i.e., (b) revelation state-based evidence construction and (c) temporally weighted evidence aggregation.
    }
    \label{fig:intro_overview}
    \vspace{-2mm}
\end{figure*}

Diffusion large language models (D-LLMs) have recently emerged as an alternative paradigm to mainstream autoregressive large language models (LLMs)~\cite{savinov2021step,li2022diffusion}. Instead of generating left-to-right one token at a time, a D-LLM iteratively denoises a token canvas and progressively reveals a subset of unresolved positions at each step (as shown in Fig.~\ref{fig:intro_overview}a)~\cite{lou2023discrete,israel2025accelerating}. Owing to this difference, D-LLMs exhibit several advantageous properties compared to their autoregressive counterparts, such as improved generation efficiency and more flexible reasoning~\cite{arriola2025block}. Despite their advantages, similar to autoregressive LLMs, D-LLMs can also suffer from hallucination issues, i.e., producing fluent and self-consistent responses that are nonetheless factually unsupported~\cite{ji2023survey,huang2025survey,pan2025survey}. This issue makes hallucination detection a critical and practical problem for improving the reliability of D-LLMs.

While the problem of hallucination detection has been widely studied since the rise of LLMs, most existing detectors are designed for autoregressive generation~\cite{duan2024shifting}, where predictions are made sequentially through a single next-token frontier. Diffusion large language models (D-LLMs), however, follow a fundamentally different decoding paradigm. In D-LLMs, tokens are generated through iterative denoising, with multiple positions being updated in parallel and gradually revealed over time. As a result, detecting hallucinations in D-LLMs requires a comprehensive understanding of both the \textit{spatial structure} of the token canvas and the \textit{temporal dynamics} of the denoising process~\cite{qian2026dynhd}. Since hallucination detection methods for autoregressive LLMs do not explicitly consider the above properties, they may not be directly applicable to D-LLMs or exhibit sub-optimal performance~\cite{chang2025tracedet}.

To fill the above gap, recent studies have explored hallucination detection for D-LLMs~\cite{chang2025tracedet,qian2026dynhd,hemmat2026tdgnet}. These pioneering studies develop well-crafted neural network-based hallucination detectors using advanced techniques, such as action trace modeling~\cite{chang2025tracedet}, deviation learning~\cite{qian2026dynhd}, and dynamic graph modeling~\cite{hemmat2026tdgnet}. Although these methods demonstrate strong detection performance compared to detection approaches for autoregressive LLMs, they require data-driven training for the detectors (i.e., training-based methods), which introduces several limitations: \ding{182}~\textbf{Limited Generalizability}. The performance of training-based methods relies heavily on dataset-specific training on the target domain. As a result, when transferred across datasets or domains, the detectors may fail to generalize effectively. \ding{183}~\textbf{Training and Deployment Cost}. Training-based methods require annotated data and incur heavy computational overhead during training, which increases both development cost and deployment complexity. \ding{184}~\textbf{Inference Efficiency}. Training-based methods often require additional detector networks, resulting in higher detection latency and reduced practicality for real-time deployment. Given these limitations, a natural question arises: 

\begin{center}
\vspace{-2mm}
\textbf{\textit{Can we design a training-free metric for hallucination detection in D-LLMs?}}
\vspace{-2mm}
\end{center}

As prediction uncertainty (typically quantified by token-level entropy) serves as a reliable indicator for hallucination detection in both autoregressive LLMs~\cite{manakul2023selfcheckgpt,farquhar2024detecting} and D-LLMs~\cite{chang2025tracedet,hemmat2026tdgnet}, we believe it provides a natural basis for designing a training-free metric. However, the iterative denoising process in D-LLMs produces entropy \textit{at each time step for every token}~\cite{sahoo2024simple,nie2025large}, which requires us to identify the most indicative signals from a large set of token entropy. 
To tackle this challenge, we decompose the problem into two dimensions. From a \textit{spatial perspective}, a long sequence of tokens is involved, each associated with its own uncertainty signal during the denoising process. 
In this context, a critical question arises: \textit{\textbf{Q1} - which tokens can provide informative evidence during the denoising process?} From a \textit{temporal perspective}, the evolving trend of entropy throughout the diffusion process provides an informative signal that can be leveraged for hallucination detection~\cite{qian2026dynhd}. 
Once the key tokens are identified at each step, a follow-up question arises: \textit{\textbf{Q2} - how should we aggregate the temporal uncertainty signals to construct a reliable indicator for hallucination detection?}

Guided by the above questions, in this paper, we propose  Temporal-weighted Revealing Entropy (\ourmethod for short), a training-free and single-run metric for hallucination detection in D-LLMs. 
We answer these questions in two stages, leveraging empirical observations and analytical reasoning to guide the design of \ourmethod. 
Specifically, to answer \textbf{\textit{Q1}}, we categorize tokens into three groups according to their revelation states at each denoising step, and our empirical analysis shows that the entropy of revealing tokens corresponding to actively generated content provides the most informative signal. Motivated by this finding, we collect the entropy mass at each diffusion step to construct the fundamental entropy-based evidence for hallucination detection (Fig.~\ref{fig:intro_overview}b). 
Then, to address \textbf{\textit{Q2}}, we analyze the temporal evolution of the evidence during the denoising process, and find that later diffusion steps contribute more than earlier ones to hallucination detection. This inspires us to design a temporal weighting scheme to aggregate the revealing-token evidence across diffusion steps to compute the final metric (Fig.~\ref{fig:intro_overview}c).
To sum up, our contributions are summarized as follows:
\begin{itemize}[leftmargin=1.2em]
    \item To the best of our knowledge, we take the first step toward a training-free hallucination detection method specifically designed for D-LLMs, which mitigates the data dependency and generalizability limitations of existing training-based detectors.
    \item We propose \ourmethod, a training-free and single-run metric for hallucination detection of D-LLMs, with designs grounded in empirical analysis. The proposed metric \ourmethod enjoys \textit{simplicity}, \textit{generality}, and \textit{accessibility}, without requiring data-driven training or multi-run sampling.
    \item Extensive experiments on multiple datasets and backbone D-LLMs demonstrate that \ourmethod achieves strong \textit{effectiveness}, \textit{computational efficiency}, \textit{robustness}, and \textit{generalizability} across diverse scenarios, highlighting its potential for practical deployment.
\end{itemize}

%% file: sections/related_work.tex
\section{Related Work}
\label{sec:related}

\noindent\textbf{Hallucination Detection in Autoregressive LLMs}
has been widely studied as a way to assess answer reliability~\cite{zhang2025siren,ji2023survey,huang2025survey}. Among them, a branch termed \textit{\textbf{training-free methods}} mainly relies on uncertainty or consistency signals~\cite{manakul2023selfcheckgpt,farquhar2024detecting,zhang2023enhancing,chen2025uncertainty}. Uncertainty-based methods use decoding-time quantities such as token likelihood, perplexity, predictive entropy, or length-normalized entropy to estimate answer reliability~\cite{kadavath2022language,zhang2023enhancing,kuhn2023semantic,farquhar2024detecting}. Differently, consistency-based methods compare multiple generations from the same input, using self-checking, lexical or semantic similarity, and semantic level uncertainty to detect unstable or contradictory responses~\cite{manakul2023selfcheckgpt,lin2023generating,chen2024inside}. These methods avoid training an additional detector, but they face an efficiency-evidence trade-off: single-run uncertainty scores are efficient but provide only limited evidence from one decoding path, whereas consistency-based scores provide richer evidence but require repeated generation, making detection substantially slower~\cite{kossen2024semantic,bi2026patchfusionmlp}. 
Another branch of hallucination detection, namely \textbf{\textit{training-based methods}}, aims to learn deep model-based hallucination detectors from richer signals, such as hidden states, latent truthfulness directions, representation-space consistency, attention features, or supervised labels~\cite{azaria2023internal,chen2024inside,zhang2024truthx,sriramanan2024llm,zhang2025prompt}. They can improve detection performance, but often require labeled data, calibration sets, auxiliary objectives, or model-specific feature engineering~\cite{burns2022discovering,zou2023representation}. Meanwhile, learned detectors are often dataset- or domain-specific, which limits their generalizability~\cite{li2023halueval,min2023factscore,dziri2022faithdial}. Although the above methods have shown effectiveness for AR-LLMs, they may not directly transfer to hallucination detection in D-LLMs, due to fundamental differences in their generation processes~\cite{sahoo2024simple,nie2025large,guo2026lost}.

\textbf{Hallucination Detection in Diffusion LLMs (D-LLMs)} is a new research direction and has been rarely explored. 
D-LLMs generate text through iterative denoising, producing uncertainty across both denoising steps and token positions, making hallucination detection more challenging~\cite{sahoo2024simple,nie2025large}. 
Existing hallucination detection methods for D-LLMs are mainly training-based, where a detector network is trained to extract informative signals from the uncertainty at denoising steps and token positions. 
For example, TraceDet learns informative entropy sub-traces from the action trace~\cite{chang2025tracedet}, while DynHD detects hallucination based on the deviations between the predicted and real entropy trajectories~\cite{qian2026dynhd}. TDGNet builds temporal dynamic graphs over evolving token-level structures to model spatiotemporal dependencies of uncertainty~\cite{hemmat2026tdgnet}. These methods show the importance of capturing token spatial and diffusion temporal signals; however, their reliance on trained detector models limits their flexibility, running efficiency, and leads to poor generalizability~\cite{varshney2022investigating,chen2024inside,li2023halueval}. In contrast, the proposed approach \ourmethod is a training-free method for hallucination detection in D-LLMs, without requiring detector training or repeated sampling. The training-free property of \ourmethod allows it to be directly applied to new application domains in a fast and efficient manner.

%% file: sections/preliminaries.tex
\section{Preliminaries}
\label{sec:preliminaries}

\textbf{Diffusion Large Language Models (D-LLMs)} generate sequences through an iterative denoising process~\cite{nie2025large,austin2021structured}. Unlike autoregressive LLMs that decode tokens from left to right, D-LLMs start from a highly masked sequence and progressively reveal tokens at different positions. Formally, given a prompt $\mathbf{q}$, the model maintains a discrete sequence state $\mathbf{x}^{(t)}=(x^{(1)}_t,\dots,x^{(l)}_t)\in\mathcal{V}^l$ at denoising step $t\in\{0,\dots,T\}$, where $\mathcal{V}$ is the vocabulary, $l$ is the fixed sequence length, and $T$ is the total number of denoising steps~\cite{austin2021structured,hoogeboom2021argmax}. The process starts from $\mathbf{x}^{(0)}$ and gradually denoises it into the final response $\mathbf{r}=\mathbf{x}^{(T)}$. At each step $t$, the model predicts a categorical distribution $\pi_i^{(t)}$ over the vocabulary for each position $i$, and we quantify the corresponding uncertainty by token entropy:
\begin{equation}
    H_t(i) = -\sum_{v\in\mathcal{V}} \pi_i^{(t)}(v)\log \pi_i^{(t)}(v).
\end{equation}
The token entropies collected across positions and denoising steps form an uncertainty trajectory, which provides the basic signal for hallucination detection~\cite{farquhar2024detecting}. 

\textbf{Hallucination Detection} aims to identify whether a generated response contains factually incorrect or unsupported content~\cite{ji2023survey}. We study hallucination detection for a question-response pair $(\mathbf{q},\mathbf{r})$, where $\mathbf{r}$ is generated by a D-LLM. Each response is assigned a binary label $y\in\{0,1\}$, where $y=1$ indicates a hallucinated response and $y=0$ denotes a factually correct one. Given a dataset $\mathcal{D}=\{(\mathbf{q}_n,\mathbf{r}_n,y_n)\}_{n=1}^{N}$, our goal is to compute a hallucination score $s(\mathbf{q},\mathbf{r})$ from the uncertainty signals observed in a single denoising trajectory. A larger score indicates a higher hallucination risk. In this work, we focus on \textbf{\textit{training-free hallucination detection}} for D-LLMs, where the score is directly constructed from entropy signals without fitting an additional detector.

%% file: sections/method.tex
\section{\ourmethod for D-LLM Hallucination Detection}
\label{sec:method}

In this section, we introduce Temporal-weighted Revealing Entropy (\ourmethod), a training-free and single-run metric for hallucination detection of D-LLMs. 
From a \textit{token-level spatial perspective}, we categorize tokens into three groups according to their revelation states at each denoising step, and then identify the token most indicative of hallucination as the fundamental entropy evidence for \ourmethod (Sec.~\ref{sec:method_revealed}). From a \textit{token-level temporal perspective}, we analyze the temporal evolution of the evidence during the denoising process, which guides the design of \ourmethod that aggregates signals along the temporal dimension (Sec.~\ref{sec:method_traj}). To better understand the mechanism of \ourmethod, we model the generation process of D-LLMs as a \textit{constraint-driven dynamical system}, from which we can interpret the key design of \ourmethod from a statistical physics perspective (Sec.~\ref{sec:physics}).

\subsection{Revelation State-Based Evidence Construction}
\label{sec:method_revealed}

Unlike autoregressive decoding, diffusion decoding does not expose a single left-to-right decision frontier~\cite{brown2020language,radford2019language}. Instead, it generates multiple tokens over arbitrary positions at each denoising step, and as the diffusion process unfolds, more tokens are gradually revealed until the final sequence is fully determined. At each diffusion step $t$, all tokens are associated with an entropy value that reflects their uncertainty; however, not all tokens are equally informative for hallucination detection. To facilitate effective hallucination detection, the key is to select the tokens with hallucination-related uncertainty signals as evidence.

\noindent\textbf{Revelation State-Based Token Categorization.} 
To identify hallucination-related tokens, we need to distinguish tokens with different roles. While tokens can be characterized by their position, semantic meaning, or uncertainty, we find that their \textit{revelation state} serves as a more effective indicator for hallucination detection. This is grounded in the inherent diffusion generation process of D-LLMs: at the initial step ($t=0$), all tokens are unrevealed and represented as masked placeholders; at each intermediate step, a certain number of tokens are ``revealing'', transitioning from an unknown masked state to a deterministic token state; at the final step ($t=T$), all tokens are fully determined. Based on their revelation states, at each diffusion step, tokens in the generated sequence can be divided into three categories: unrevealed tokens, revealing tokens, and revealed tokens. 

Formally, at diffusion step $t$, the unrevealed token set is denoted as $\mathcal{T}^u_{t}$, where each token is masked and remains in an uncertain state; the revealing token set is denoted as $\mathcal{T}^g_{t}$, where tokens are transitioning from the masked state to a determined state at the current step; and the revealed token set is denoted as $\mathcal{T}^r_{t}$, where tokens have been determined in previous steps. There is no overlap between the three sets, and their union $\mathcal{T}^u_t \cup \mathcal{T}^g_t \cup \mathcal{T}^r_t = \mathcal{T}$ forms the complete token set. Taking the second step $t=2$ in Fig.~\ref{fig:intro_overview}a as an example, the revealed token is ``Prejudice'', the revealing token is ``wrote'', while the remaining tokens are unrevealed.

The rationale behind such categorization is that the revelation state reflects the certainty level of each token at each step, and thus links its entropy to its role in hallucination detection. Specifically, the entropy of unrevealed tokens may reflect the model’s uncertainty over a large space of possible outcomes. In contrast, the entropy of revealed tokens is typically less informative, as it only reflects residual uncertainty around already determined tokens.
Furthermore, the entropy of revealing tokens directly corresponds to the uncertainty of the currently generating tokens, and thus directly reflects the confidence of the generation. Considering that the entropy of the three types of tokens plays different roles, we posit that their contributions to hallucination detection can be reasonably differentiated.

\noindent\textbf{Empirical Analysis.}
To understand the roles of different token groups, we empirically investigate how their entropy contributes to hallucination detection. Specifically, for each token type, we first compute the average entropy for each sample, and then use Cohen’s d to quantify the difference in entropy between hallucinated and non-hallucinated samples. 
We use late-stage entropy (i.e., from the last 30\% of denoising steps) as evidence, since later steps tend to be more informative (see Sec.~\ref{sec:method_traj} for detailed analysis), and employ mean aggregation to summarize these late-stage signals. 
More concretely, the raw evidence in this experiment can be written as: 
\begin{wrapfigure}{r}{0.53\textwidth}
\vspace{-3mm} 
\centering
\includegraphics[width=0.53\textwidth]{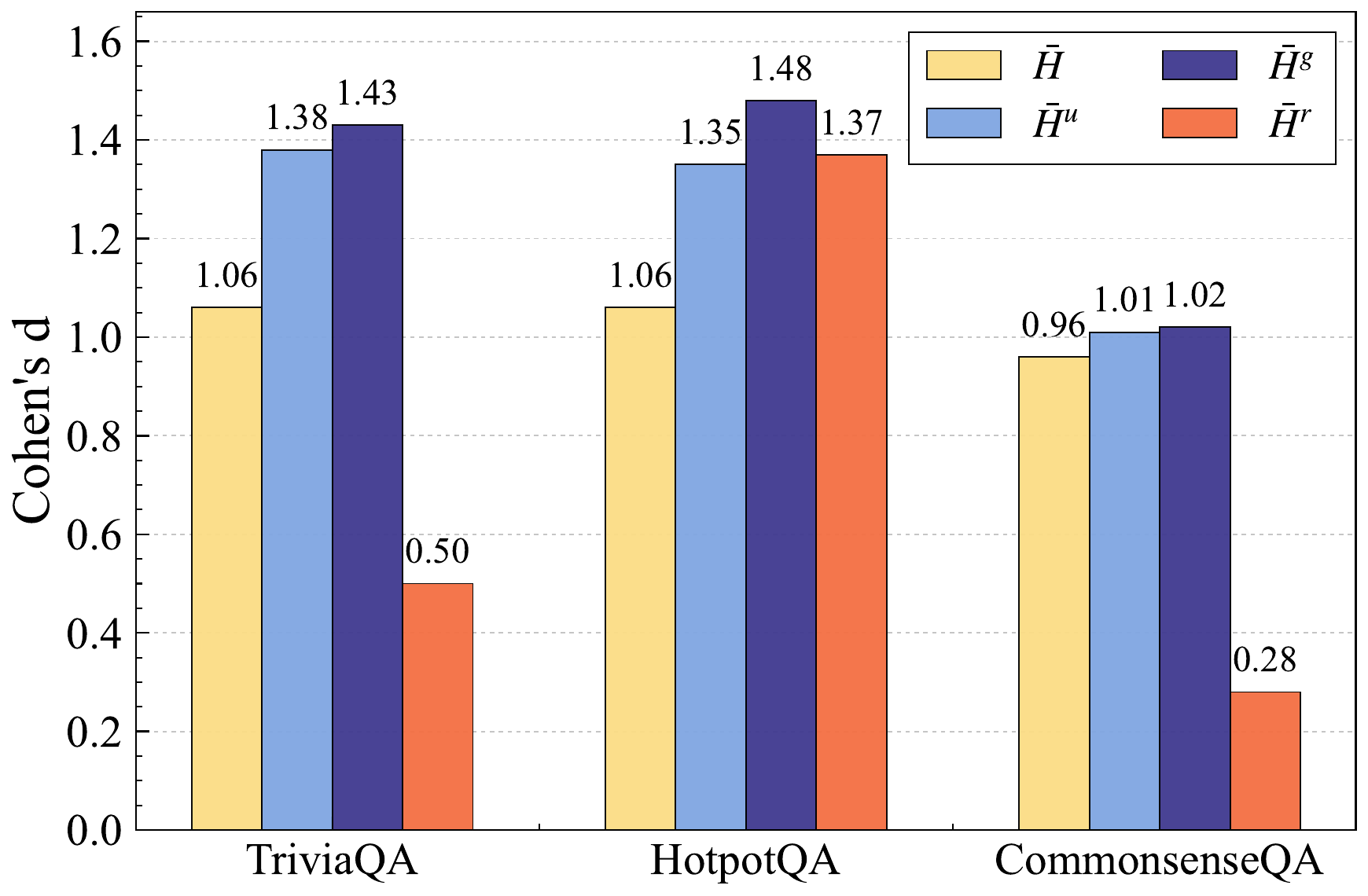}\vspace{-2mm}
\caption{Evidence informativeness comparison on LLaDA-8B-Instruct with 64 diffusion steps.}
\label{fig:moti_evidence}
\vspace{-8mm}
\end{wrapfigure}

\begin{equation}
\bar{H}^{c}
:= \frac{1}{|\mathcal{S}|}
\sum_{t \in \mathcal{S}}
\frac{1}{|\mathcal{T}^{c}_{t}|}
\sum_{i \in \mathcal{T}^{c}_{t}} H_t(i),
\end{equation}

\noindent where $\mathcal{S}$ denotes the set of late diffusion steps, 
$\mathcal{T}^{c}_{t}$ denotes the token set of category $c$ at step $t$, 
and $H_t(i)$ denotes the entropy of token $i$ at diffusion step $t$, with category $c \in \{u,g,r\}$
and $u$, $g$, and $r$ corresponding to unrevealed, revealing, and revealed tokens, respectively. 

Taking $\bar{H}^{u}$, $\bar{H}^{g}$, and $\bar{H}^{r}$ as evidence, we compute Cohen’s d to quantify the difference in these values between hallucinated and non-hallucinated samples. A larger Cohen’s d indicates a stronger separability between the two groups, and thus better discriminative power for hallucination detection. We also compute the average late-stage entropy over all tokens, $\bar{H}$, as an additional baseline. According to the results shown in Fig.~\ref{fig:moti_evidence} (more results are in Appendix~\ref{app:evidence_construction}), we have the following observations. 
\ding{182}~$\bar{H}^g$ consistently achieves the best discriminateness, indicating that the entropy of revealing tokens is most aligned with the uncertainty of the ongoing generation process and thus most relevant to hallucination detection. 
\ding{183}~$\bar{H}^r$ shows limited discriminative power, as it mainly reflects residual uncertainty of already determined tokens. 
\ding{184}~The overall entropy $\bar{H}$ performs moderately, suggesting that aggregating all tokens may dilute informative signals. 
\ding{185}~$\bar{H}^u$ remains reasonably discriminative, as it captures uncertainty over yet-to-be-determined tokens. 
An in-depth interpretation of $\bar{H}^g$'s behavior is provided in Appendix~\ref{app:evidence_construction}.

\noindent\textbf{Entropy Mass as Evidence Construction.}
As the entropy of revealing tokens consistently shows strong discriminative power for hallucination detection, we employ ${H}^{g}_t$ as the primary evidence for \ourmethod. Specifically, we collect the entropy of revealing tokens at all diffusion steps, forming an evidence sequence $\mathbf{e}\in \mathbb{R}^{T}$ for hallucination detection:
\begin{equation}
\mathbf{e}
= \{\tilde{H}_t^{g}\}_{t=1}^{T}
= \{ 
\sum_{i \in \mathcal{T}_t^{g}} H_t(i)\}_{t=1}^{T}.
\end{equation}

\noindent Note that we use the mass entropy $\tilde{H}_t^{g}$, rather than the average entropy $\bar{H}_t^{g}$ as evidence. This is because hallucination risk depends not only on the uncertainty of each revealing token, but also on the amount of uncertain content being revealed. 
The entropy mass can capture both the intensity of uncertainty and the volume of revealed tokens, which can be interpreted as the entropy flux across the evolving reveal boundary at step $t$. Therefore, we argue that average entropy may lose information about the volume of uncertainty, whereas entropy mass provides a more informative signal. 
Also, we do not include $t=0$, since all tokens are still masked at this stage, yielding non-informative entropy. The revealing token-based evidence then provides the raw signal for computing our training-free hallucination detection metric \ourmethod.

\subsection{Temporally Weighted Evidence Aggregation}
\label{sec:method_traj}

Although the above empirical results show that revealing token-based evidence can provide useful signals for hallucination detection, mean aggregation across different diffusion steps ignores the temporal variation of these signals. Indeed, in diffusion models, different denoising steps typically play different roles in the generation process. For example, in diffusion models for image generation, early steps tend to capture coarse structures, while later steps focus on refining fine-grained details~\cite{ho2020denoising,dhariwal2021diffusion,luo2022understanding}. In recent studies of training-based D-LLM hallucination detection, researchers have also observed that signals from different denoising steps exhibit varying levels of discriminative power~\cite{qian2026dynhd,chang2025tracedet}. Nevertheless, unlike training-based approaches, where the importance of different denoising steps can be automatically learned, in our training-free setting, the contribution of each step needs to be specified through a designed aggregation scheme. To effectively aggregate the evidence, a natural question is: \textit{Which time steps contribute more to hallucination detection?} 

\noindent\textbf{Empirical Analysis.}
To address the above question, we analyze the evidence sequences of both hallucinated and non-hallucinated samples to uncover their temporal patterns. Taking the entropy mass as evidence, we visualize the evidence sequences of different D-LLMs (LLaDA and Dream)~\cite{nie2025large,ye2025dream}, datasets (TriviaQA and HotpotQA)~\cite{yang2018hotpotqa,joshi2017triviaqa}, and total diffusion steps (64 and 128), to derive a comprehensive and generalizable pattern. 

\begin{figure*}[t]
\vspace{-2mm}
    \centering
    \begin{subfigure}[t]{0.24\textwidth}
        \centering
        \includegraphics[height=2.5cm]{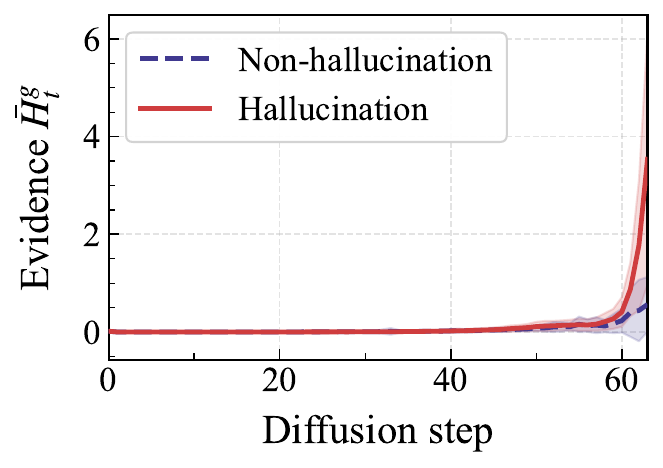}
        \caption{HQA-LLaDA(64)}
        \label{fig:hotpot64}
    \end{subfigure}
    \hfill
    \begin{subfigure}[t]{0.225\textwidth}
        \centering
        \includegraphics[height=2.5cm]{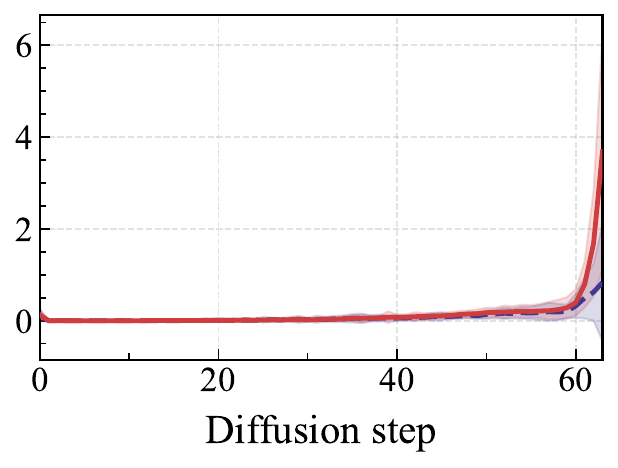}
        \caption{TQA-LLaDA(64)}
        \label{fig:tqa64}
    \end{subfigure}
    \hfill
    \begin{subfigure}[t]{0.225\textwidth}
        \centering
        \includegraphics[height=2.5cm]{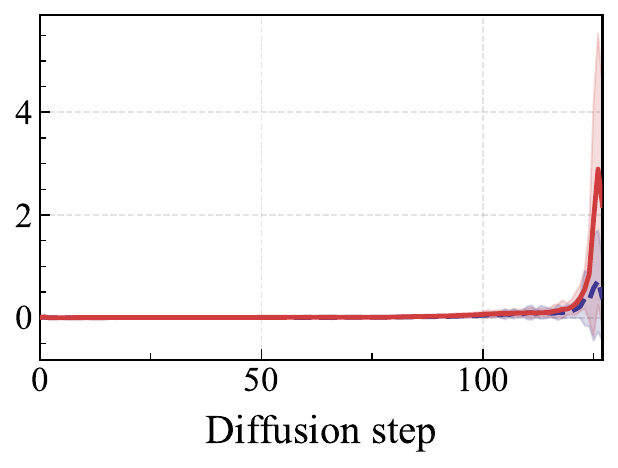}
        \caption{HQA-LLaDA(128)}
        \label{fig:hotpot128}
    \end{subfigure}
    \hfill
    \begin{subfigure}[t]{0.225\textwidth}
        \centering
        \includegraphics[height=2.5cm]{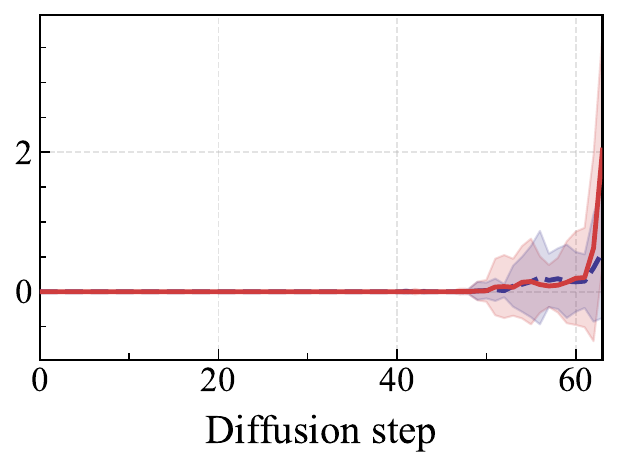}
        \caption{HQA-Dream(64)}
        \label{fig:hotpot64_dream}
    \end{subfigure}
\vspace{-2mm}
    \caption{
    Revealing-token entropy trajectories across different D-LLMs, datasets, and diffusion steps.
    }
    \vspace{-5mm}
    \label{fig:revealed_traj_main}
\end{figure*}

The visualization results are shown in Fig.~\ref{fig:revealed_traj_main} (more results are in Appendix~\ref{app:revealing_trajectory}), which lead to the following findings. \ding{182}~Across all four settings, the gap between hallucination and non-hallucination curves remains modest early in denoising but widens sharply near the end. This suggests a general pattern that \textbf{later diffusion steps} are more informative for hallucination detection. \ding{183}~For both hallucinated and non-hallucinated samples, the entropy remains nearly constant with negligible differences during early denoising steps, resulting in flat trajectories. This indicates that, at early stages of generation, the model primarily focuses on capturing coarse or global structures, rather than content that may lead to hallucination. \ding{184}~At later diffusion steps, the evidence trajectories of the two groups diverge markedly. Specifically, hallucinated samples exhibit a sharp upward trend, with values rising from around 0 to approximately 2$\sim$3; in contrast, non-hallucinated samples show only modest increases and may even fluctuate in some cases.

\noindent\textbf{Discussion: Why Later Steps Matter?}
To explain the above phenomenon, we provide the following interpretation and analysis. 
\textit{First}, hallucinations often emerge when the model commits to specific factual details, such as entities, relations, or attributes~\cite{maynez2020faithfulness}. 
These fine-grained decisions are more likely to be finalized in later steps, since the denoising process progressively refines tokens from uncertain placeholders to deterministic outputs. 
\textit{Second}, later steps accumulate the effects of previous denoising decisions, so uncertainty at this stage reflects not only local token ambiguity but also the consistency of the partially generated sequence. 
Therefore, late-stage evidence better captures whether the generation trajectory is converging toward a reliable answer or drifting toward hallucination. 
\textit{Third}, early steps mainly capture coarse structure, where predictions remain ambiguous and are less tied to factual correctness.

\paragraph{\ourmethod Metric Design.}
Based on the above empirical observations and analysis, we conclude that entropy-based evidence at later stages should be emphasized compared to earlier stages. However, to develop a general and robust training-free metric, it is non-trivial to define a fixed threshold to determine the onset of the late stage, as this point may vary across different models, datasets, and diffusion settings. To address this issue, instead of defining a threshold, we adopt a monotonic weighting schedule to reweight the evidence. Concretely, we assign a weight to each time step, where earlier steps receive smaller weights and later steps receive larger weights. 
Among all monotone schedules, we adopt the simplest parameter-free rule, $w_t = \frac{t}{T}$, where the weight increases linearly with the diffusion step. 
Applying this monotonic weighting, the final score of \ourmethod\ can be defined as:
\begin{equation}
\ourmethod:= \sum_{t=1}^{T} w_t \, \tilde{H}_t^{g}
= \sum_{t=1}^{T} \frac{t}{T} \sum_{i \in \mathcal{T}_t^{g}} H_t(i).
\label{eq:rcr_final}
\end{equation}


Compared to existing hallucination detection methods, \ourmethod offers the following merits. 
\ding{182}~\textbf{Simplicity.} It requires only a single run and does not rely on external knowledge verification or additional LLM calls, making it lightweight and well-suited for real-time applications. 
\ding{183}~\textbf{Generality.} As a training-free approach, it can be directly applied without training, avoiding potential biases introduced by data or model fitting. 
\ding{184}~\textbf{Accessibility.} It relies solely on token-level entropy and does not require access to internal model states, making it applicable in gray-box settings where model parameters or internal states are not accessible. 
\ding{185}~\textbf{D-LLM Compatibility.} It is specifically designed for D-LLMs by leveraging the denoising process and token revelation dynamics, making it better suited to D-LLMs compared to parameter-free methods designed for AR-LLMs. 

%% file: sections/method33.tex
\subsection{Understanding \ourmethod from a Statistical Physics Perspective} \label{sec:physics}

\begin{wrapfigure}{r}{0.53\textwidth}
\vspace{-7mm} 
\centering
\includegraphics[width=0.53\textwidth]{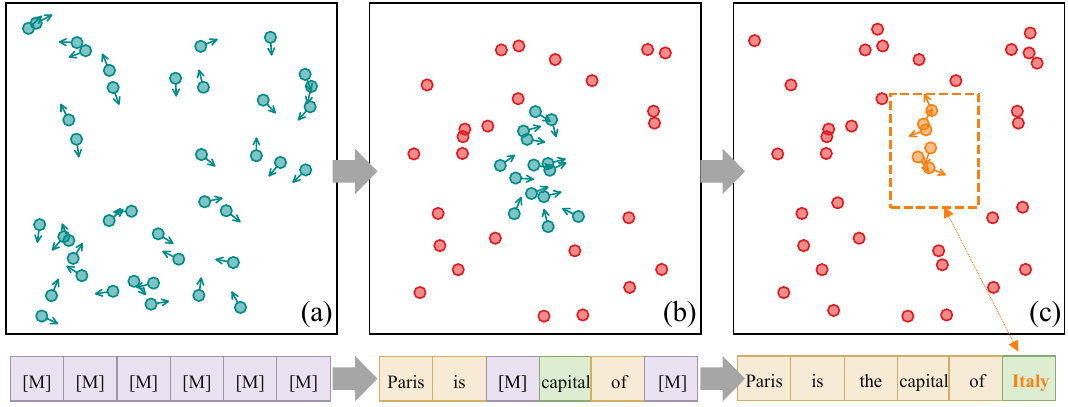}\vspace{-1mm}
\caption{Analogy of CDSs and D-LLMs.}
\label{fig:moti_cds}
\vspace{-7mm}
\end{wrapfigure}
While empirical results show that \ourmethod is effective, it is natural to ask whether there exists an intuitive perspective to interpret its underlying mechanism. Motivated by this, we draw an analogy between the revealing process of D-LLMs and that of a constrained dynamical system (CDS), and then use this perspective to interpret the design of \ourmethod. 

In statistical physics, CDSs (e.g., spin glass models) refer to systems whose evolution is restricted by state or interaction constraints over time~\cite{khalil2002nonlinear,posa2016optimization}. 
The evolution of a CDS provides a natural analogy to the revealing process of D-LLMs, where variables (tokens) are gradually fixed while the remaining degrees of freedom are updated. 
Specifically, as shown in Fig.~\ref{fig:moti_cds}, \textit{at the initial stage}, a CDS is only weakly constrained, and uncertainty is relatively uniformly distributed across the system. Similarly, in the early stage of D-LLM decoding, no tokens are determined, and entropy is spread more evenly across token positions. 
\textit{As time progresses}, additional constraints are imposed, which restricts part of the state space. The remaining variables must adapt within a reduced feasible region, leading to a redistribution of uncertainty. In D-LLMs, this corresponds to revealing a subset of tokens, after which the remaining tokens carry a larger share of the residual uncertainty, resulting in entropy concentration. 
\textit{In some late steps of the dynamical process}, especially under incompatible or conflicting constraints, e.g., conflicting boundary conditions~\cite{mezard1988spin} or inconsistent local interactions~\cite{binder1986spin}, the system may enter a frustrated state where no globally consistent configuration exists. This leads to localized fluctuations and instability. In D-LLM decoding, such localized uncertainty amplification can be the main cause of increased conditional entropy at certain positions, thereby increasing the likelihood of hallucination. \looseness=-1

Under this analogy, we can better understand the \textit{design intuition} of \ourmethod. 
\ding{182}~\textbf{Why revealing entropy is informative?} (Details in Appendix~\ref{app:selection_boundary_flux}) Revealing tokens lie at the moving boundary where uncertainty is committed into the answer, so their entropy directly measures the amount of uncertain content being fixed. 
\ding{183}~\textbf{Why late-stage entropy matters?} (Details in Appendix~\ref{app:boundary_flux_to_tre}) As constraints accumulate, uncertainty is progressively concentrated on fewer remaining degrees of freedom, making late-stage entropy more informative about the final outcome. 
\ding{184}~\textbf{Why hallucination samples have higher late-stage entropy?} (Details in Appendix~\ref{app:attention_boundary_conditions} and~\ref{app:boundary_susceptibility}) Incompatible commitments create frustrated boundary conditions, which destabilize the remaining subsystem and amplify uncertainty at later stages. 

Notably, while this analogy provides an intuitive explanation for the decoding behavior of D-LLMs and the design of \ourmethod, a rigorous theoretical characterization remains beyond the scope of this work. A more detailed discussion is provided in Appendix~\ref{app:cds}, and we hope this can offer inspiration for future theoretical understanding of D-LLMs.

%% file: sections/experiments.tex
\section{Experiments}
\label{sec:exp}

\subsection{Experimental Setup}
\label{subsec:setup}

\noindent\textbf{Baselines.} We compare \ourmethod with training-based detectors, including CCS~\cite{burns2022discovering}, TSV~\cite{park2025steer}, and TraceDet~\cite{chang2025tracedet}, and also training-free methods, including Perplexity~\cite{ren2022out}, LN-Entropy~\cite{malinin2020uncertainty}, Semantic Entropy~\cite{kuhn2023semantic}, Lexical Similarity~\cite{lin2023generating}, and EigenScore~\cite{chen2024inside}. Note that TraceDet is the state-of-the-art training-based method for D-LLMs. We would not compare with~\cite{qian2026dynhd,hemmat2026tdgnet} due to the lack of open-source code. 

\noindent\textbf{Evaluation Protocol.} Following~\cite{chang2025tracedet}, we evaluate on two D-LLM families, \llada-8B-Instruct~\cite{nie2025large} and \dream-7B-Instruct~\cite{ye2025dream}, on three QA benchmarks:
TriviaQA~\cite{joshi2017triviaqa}, HotpotQA~\cite{yang2018hotpotqa}, and CommonsenseQA~\cite{talmor2019commonsenseqa}. We use AUROC for evaluation. The training-based detectors are trained on the corresponding datasets with automatic annotations, following~\cite{chang2025tracedet}. The training-free methods are applied directly  without optimization. For Dream-7B-Instruct, we omit Perplexity and LN-Entropy due to restricted access to stable token-level logits. More experimental details are in Appendix~\ref{app:exp_setup}.

\input{tables/main_results}

\subsection{Results and Analysis}
\label{subsec:results}

\noindent\textbf{Main Comparison.}

\begin{wraptable}[11]{r}{0.40\textwidth}
\vspace{-4mm}
\centering
\caption{Ablation study on \llada.}
\vspace{-1mm}
\label{tab:ablation}
\resizebox{0.42\textwidth}{!}{
\begin{tabular}{l|ccc}
\toprule
Variant & TriQA & HotQA & CSQA \\
\midrule
All tokens $\bar H$ & 64.0 & 70.9 & 75.7 \\
Unrevealed $\bar H^u$ & 63.9 & 70.8 & 75.9 \\
Revealed $\bar H^r$ & 53.9 & 84.0 & 60.2 \\
\midrule
Average & 81.1 & 85.1 & 79.0 \\
Exponential & 82.0 & 85.6 & 79.0 \\
Hard last-$30\%$ & \textbf{82.4} & 85.4 & {79.0} \\
\midrule
\ourmethod (ours) & 82.2 & \textbf{85.6} & \textbf{79.0} \\
\bottomrule
\end{tabular}
}
\vspace{-3mm}
\end{wraptable}

Table~\ref{tab:main} reports the comparison results, from which we draw the following observations. \ding{182}~\ourmethod is consistently competitive across both D-LLM families and three QA datasets while requiring no training or additional generation. The gain is especially clear on \llada, while \ourmethod's performance is also competitive on \dream.
\noindent\ding{183}~Detection approaches specifically designed for D-LLMs (i.e., TraceDet and \ourmethod) significantly outperform those developed for autoregressive models, highlighting the importance of accounting for the diffusion-based generation mechanism. 
\ding{184}~\ourmethod shows comparable or even better performance compared to TraceDet, a training-based D-LLM hallucination detection method. This indicates that our training-free method can achieve strong performance without requiring dedicated detector training.

\noindent\textbf{Ablation Study.}
To investigate the contribution of each design, we replace the evidence source and weighting scheme in \ourmethod, and the results in Table~\ref{tab:ablation} (more results are in Appendix~\ref{app:additional_ablation}) lead to the following findings. \ding{182}~Using the entropy of other token groups leads to a significant performance drop, indicating that revealing tokens provides the most informative signals. \ding{183}~Average weighting leads to sub-optimal performance. While exponential weighting and threshold-based selection may sometimes yield slight improvements, they introduce additional hyperparameters (e.g., exponent/threshold), so we adopt a simple parameter-free linear weighting scheme. \ding{184}~\ourmethod generally achieves competitive performance, indicating that modeling revealing-token entropy with temporal weighting is an effective choice for training-free hallucination detection.

\begin{figure*}[t]
\vspace{-2mm}
    \centering
    \begin{subfigure}[t]{0.32\textwidth}
        \centering
        \includegraphics[width=\linewidth]{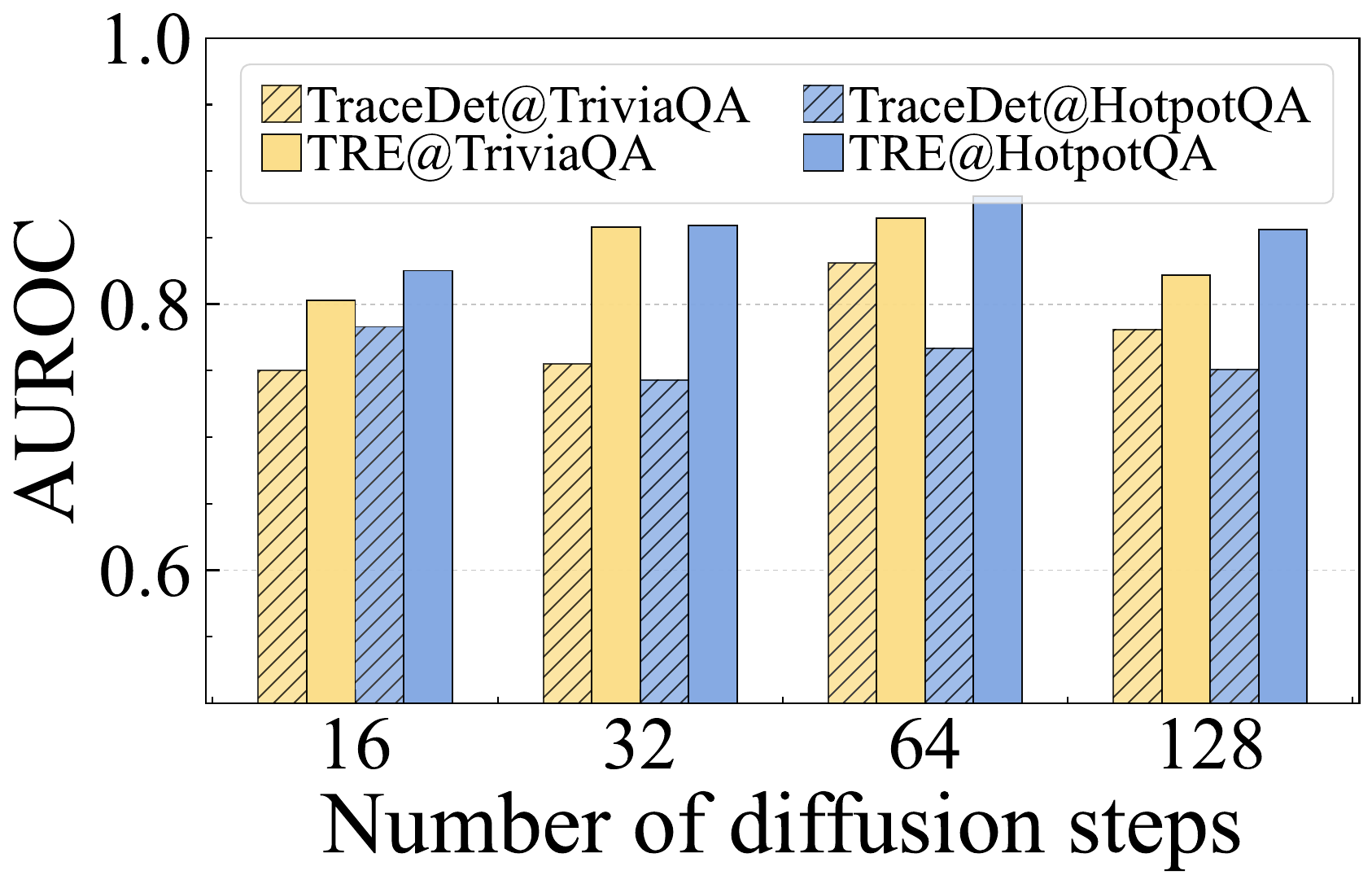}
        \caption{Diffusion Steps}
        \label{fig:sen_step}
    \end{subfigure}
    \hfill
    \begin{subfigure}[t]{0.32\textwidth}
        \centering
        \includegraphics[width=\linewidth]{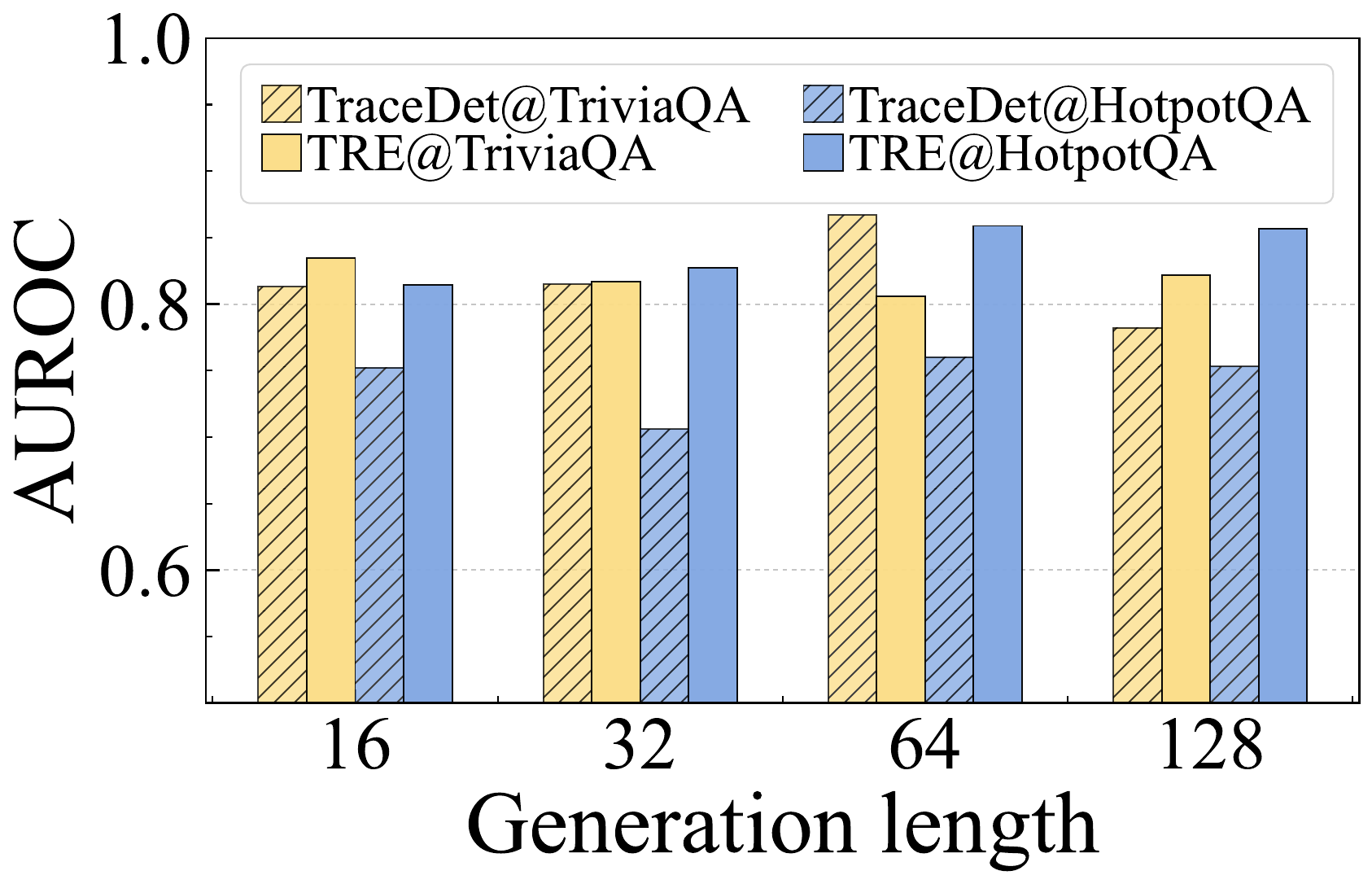}
        \caption{Generation Length}
        \label{fig:sen_length}
    \end{subfigure}
    \hfill
    \begin{subfigure}[t]{0.32\textwidth}
        \centering
        \includegraphics[width=\linewidth]{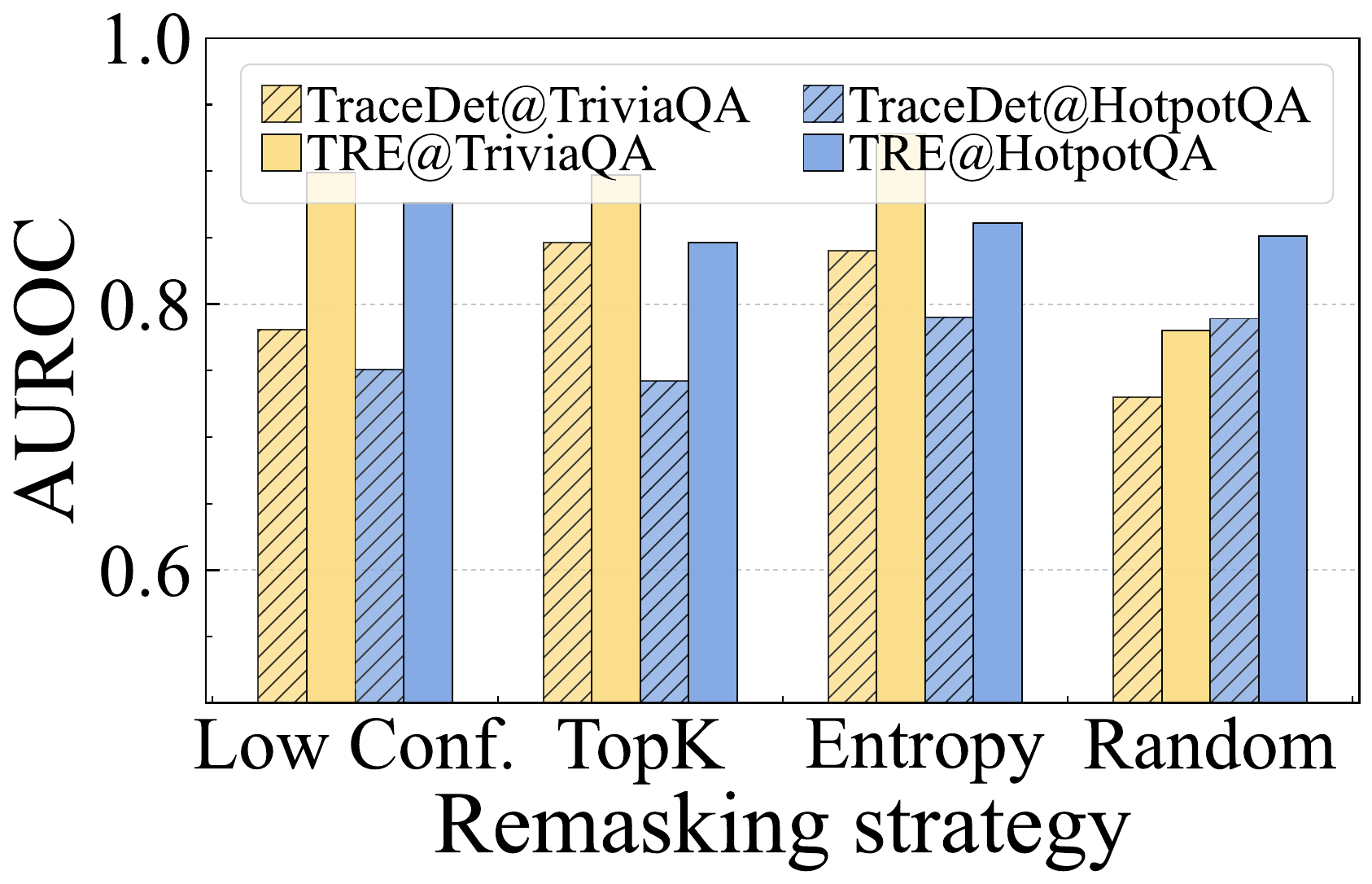}
        \caption{Remasking Strategies}
        \label{fig:sen_remask}
    \end{subfigure}
\vspace{-1mm}
    \caption{
    Sensitivity w.r.t. (a)~Diffusion Steps, (b)~Generation Length, and (c)~Remasking Strategies.
    }
    \label{fig:sen}
\vspace{-5mm}
\end{figure*}

\begin{wrapfigure}{r}{0.33\textwidth}
 \centering
 \vspace{-7mm}
  \includegraphics[width=0.33\textwidth]{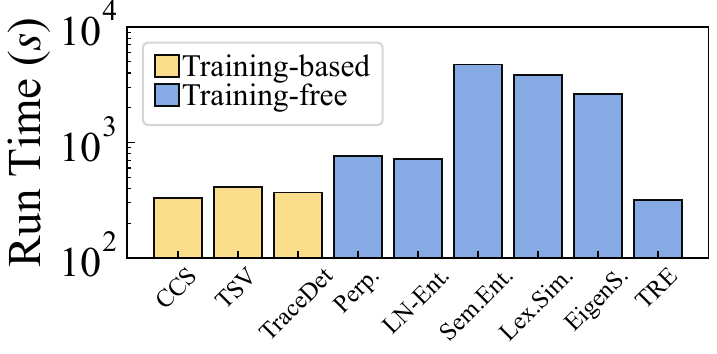}
  \vspace{-6mm}
  \caption{Time comparison.}
  \label{fig:efficency}
  \vspace{-2mm}
\end{wrapfigure}

\noindent\textbf{Efficiency Analysis.} 
To assess the runtime efficiency of \ourmethod, we compare the inference time on 100 QA samples from TriviaQA dataset using LLaDA. As depicted in Fig.~\ref{fig:efficency}, \ourmethod demonstrates the shortest runtime and significantly outperforms multi-run methods (e.g., LN-Entropy and Semantic Entropy) in terms of efficiency. Notably, training-based methods require additional time for training, which is not included in the reported runtime.

\begin{wrapfigure}{r}{0.48\textwidth}
  \centering
  \vspace{-3mm}
  \includegraphics[width=0.48\textwidth]{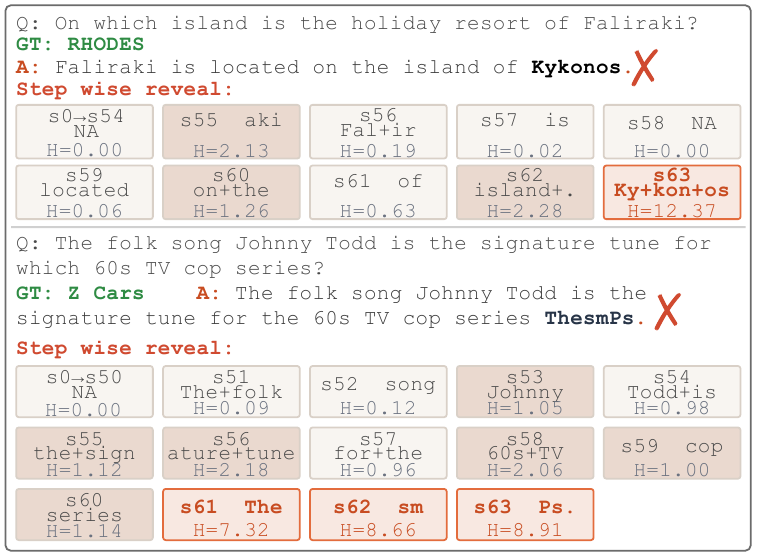}
  \caption{Case study of step-wise reveal.}
  \label{fig:case_study}
  \vspace{-3mm}
\end{wrapfigure}
\noindent\textbf{Sensitivity Analysis.}
We further study the sensitivity of \ourmethod to generation time steps, generation lengths, and remasking strategies, as shown in Fig.~\ref{fig:sen}. The results lead to the following findings. \ding{182}~\ourmethod is robust to generation time steps. Across different settings, TRE consistently outperforms TraceDet on both TriviaQA and HotpotQA, with only minor fluctuations. \ding{183}~\ourmethod is also insensitive to generation length. TRE maintains competitive AUROC under various length budgets, suggesting that revealing-token entropy provides reliable signals without carefully tuning the generation length. \ding{184}~For remasking strategies, all four decoding strategies achieve consistently strong average performance, indicating that \ourmethod remains effective across different decoding configurations.

\noindent\textbf{Case Study.}
We provide two hallucinated examples in Fig.~\ref{fig:case_study} (more can be found in Appendix~\ref{app:case_study}) to illustrate the uncertainty dynamics captured by \ourmethod. In the figure, the generated tokens at each step (s) are listed, with the revealing entropy (H) value given. Our observations are as follows: \ding{182}~Hallucinated entities are often revealed at late denoising steps, such as ``Kykonos'' and ``ThesmPs'', indicating that late-stage cues are usually critical. \ding{183}~These hallucinated tokens show a sharply increased entropy when revealed, while earlier tokens usually have a near-zero or much smaller entropy. This supports our motivation that late-stage newly revealed-token entropy serves as a direct and informative signal for hallucination detection.

%% file: tables/main_results.tex
\newcolumntype{C}{>{\centering\arraybackslash}X} 
\begin{table*}[t]
\centering
\small
\setlength{\tabcolsep}{3pt} 
\caption{AUROC(\%) comparison of hallucination detection methods on two D-LLMs across three QA datasets. The highest score is \textbf{bolded} and the second highest is \underline{underlined}.}
\label{tab:main}
\begin{tabularx}{\textwidth}{l c *{6}{C} c} 
\toprule
\multirow{2}{*}{\textbf{Method}}
& \multirow{2}{*}{\shortstack{\textbf{Designed}\\\textbf{for D-LLMs}}}
& \multicolumn{2}{c}{\textbf{TriviaQA}}
& \multicolumn{2}{c}{\textbf{HotpotQA}}
& \multicolumn{2}{c}{\textbf{CSQA}}
& \multirow{2}{*}{\textbf{Avg}} \\
\cmidrule(lr){3-4}\cmidrule(lr){5-6}\cmidrule(lr){7-8}
& & 128 & 64 & 128 & 64 & 128 & 64 & \\
\midrule

\rowcolor[HTML]{DEDEDE} \multicolumn{9}{l}{\textbf{LLaDA-8B-Instruct}} \\
\rowcolor[HTML]{F5F5F5} \multicolumn{9}{l}{\textit{Training-based Methods}} \\
 \quad CCS  &      \ding{55}        & 57.1 & 54.2 & 57.6 & 55.8 & 50.5 & 58.5 & 55.6 \\
 \quad TSV  &      \ding{55}        & 60.2 & 61.1 & 65.0 & 59.4 & 52.9 & 55.2 & 59.0 \\
 \quad TraceDet   &    \ding{51}    & \underline{73.9} & \underline{74.1} & \underline{66.1} & \underline{63.7} & \underline{77.2} & \textbf{77.1} & \underline{72.0} \\
\rowcolor[HTML]{F5F5F5} \multicolumn{9}{l}{\textit{Training-free Methods}} \\
 \quad Perplexity &   \ding{55}     & 50.4 & 47.6 & 49.3 & 51.2 & 65.6 & 65.0 & 54.9 \\
 \quad LN-Entropy &    \ding{55}    & 54.6 & 53.5 & 54.8 & 54.7 & 64.6 & 64.4 & 57.8 \\
 \quad Semantic Entropy&  \ding{55} & 68.9 & 67.3 & 57.6 & 53.8 & 44.1 & 43.9 & 55.9 \\
 \quad Lexical Similarity& \ding{55}& 62.5 & 59.0 & 64.2 & 57.1 & 57.3 & 60.7 & 60.1 \\
 \quad EigenScore  &    \ding{55}   & 69.2 & 66.9 & 64.7 & 59.2 & 58.5 & 60.6 & 63.2 \\
 \quad \textbf{\ourmethod} (ours) & \ding{51}& \textbf{82.2} & \textbf{86.5} & \textbf{85.6} & \textbf{88.1} & \textbf{79.0} & \underline{75.0} & \textbf{82.7} \\

\midrule

\rowcolor[HTML]{DEDEDE} \multicolumn{9}{l}{\textbf{Dream-7B-Instruct}} \\
\rowcolor[HTML]{F5F5F5} \multicolumn{9}{l}{\textit{Training-based Methods}} \\
 \quad CCS  &       \ding{55}       & 56.9 & 50.3 & 51.7 & 58.2 & 54.2 & 53.2 & 54.1 \\
 \quad TSV  &      \ding{55}        & 75.6 & 74.7 & 58.7 & 63.0 & 62.3 & 56.8 & 65.2 \\
 \quad TraceDet   &   \ding{51}     & \underline{78.1} & \textbf{86.7} & \underline{75.1} & \underline{76.0} & \textbf{84.7} & \textbf{84.1} & \underline{80.8} \\
\rowcolor[HTML]{F5F5F5} \multicolumn{9}{l}{\textit{Training-free Methods}} \\
 \quad Semantic Entropy & \ding{55} & 73.7 & 72.5 & 62.7 & 67.7 & 51.4 & 48.6 & 62.8 \\
 \quad Lexical Similarity& \ding{55}& 58.3 & 64.0 & 59.7 & 62.7 & 77.3 & 76.9 & 66.5 \\
 \quad EigenScore   &   \ding{55}   & 66.0 & 69.1 & 62.5 & 67.0 & 76.9 & 77.5 & 69.8 \\
 \quad \textbf{\ourmethod} (ours) &\ding{51}& \textbf{83.9} & \underline{84.8} & \textbf{80.6} & \textbf{81.0} & \underline{77.7} & \underline{78.4} & \textbf{81.1} \\

\bottomrule
\end{tabularx}
\vspace{-4mm}
\end{table*}

%% file: sections/conclusion.tex
\section{Conclusion}
\label{sec:conclusion}

\vspace{-2mm}
In this paper, we introduce \ourmethod, a training-free and single-run hallucination detector for diffusion language models. Unlike prior detectors that rely on additional training or repeated generation, \ourmethod exploits the diffusion decoding trajectory by aggregating the uncertainty of revealing tokens with a simple monotone temporal weighting scheme, emphasizing late commitment events that are more indicative of hallucination. Extensive experiments on multiple QA datasets and diffusion language models demonstrate that \ourmethod achieves strong, efficient, and robust performance with low sensitivity to decoding configurations. 
A potential \textbf{limitation} is that \ourmethod relies solely on entropy-based signals and does not explicitly incorporate semantic or factual knowledge, which may limit its effectiveness in cases where uncertainty is poorly aligned with factual correctness.
\textbf{Future work} may explore integrating semantic-aware signals into training-free D-LLM hallucination detection framework to incorporate richer contextual consistency measures.

%% file: appendix/00_content.tex
\section{Details of Motivating Experiments}

\subsection{Evidence Construction Experiments}
\label{app:evidence_construction}

This appendix provides additional details for the evidence construction experiment in Sec.~\ref{sec:method_revealed}. 
The main text reports the comparison on \llada with 64 diffusion steps. 
Here, we provide the corresponding results with 128 diffusion steps to verify whether the revelation-state based evidence remains informative under a longer denoising trajectory~\cite{zuo2025rethinking}.

\paragraph{Token-Family Evidence.}
At each denoising step $t$, we divide token positions into three revelation-state based groups: unrevealed tokens $T_t^u$, revealing tokens $T_t^g$, and revealed tokens $T_t^r$. 
For each token family $c\in\{u,g,r\}$, we compute its late-stage mean entropy as
\begin{equation}
\bar H^c
:=
\frac{1}{|S|}
\sum_{t\in S}
\frac{1}{|T_t^c|}
\sum_{i\in T_t^c}
H_t(i),
\qquad
c\in\{u,g,r\},
\label{eq:app_late_family_entropy}
\end{equation}
where $S$ denotes the set of late diffusion steps and $H_t(i)$ is the token entropy at position $i$ and step $t$. 
We also compute the all-token late-stage entropy baseline:
\begin{equation}
\bar H
:=
\frac{1}{|S|}
\sum_{t\in S}
\frac{1}{L}
\sum_{i=1}^{L}
H_t(i).
\label{eq:app_late_all_entropy}
\end{equation}
In our experiments, $S$ is chosen as the last 30\% of denoising steps.
This is used only for evidence analysis, while the final TRE score aggregates all denoising steps through temporal weighting.

\paragraph{Cohen's $d$.}
To quantify the discriminative power of each scalar evidence, we use Cohen's $d$, a standard effect-size measure for two-group separation. 
Given a scalar evidence value $z$, let $(\mu_1,s_1,n_1)$ and $(\mu_0,s_0,n_0)$ denote the sample mean, standard deviation, and sample count for hallucinated and non-hallucinated samples, respectively. 
Cohen's $d$ is computed as
\begin{equation}
d
=
\frac{\mu_1-\mu_0}{s_{\mathrm{pooled}}},
\qquad
s_{\mathrm{pooled}}
=
\sqrt{
\frac{(n_1-1)s_1^2+(n_0-1)s_0^2}{n_1+n_0-2}
}.
\label{eq:app_cohens_d}
\end{equation}
A larger positive value indicates that the evidence is higher on hallucinated samples and better separated relative to within-group variability.

\begin{figure}[b]
    \centering
    \includegraphics[width=0.55\linewidth]{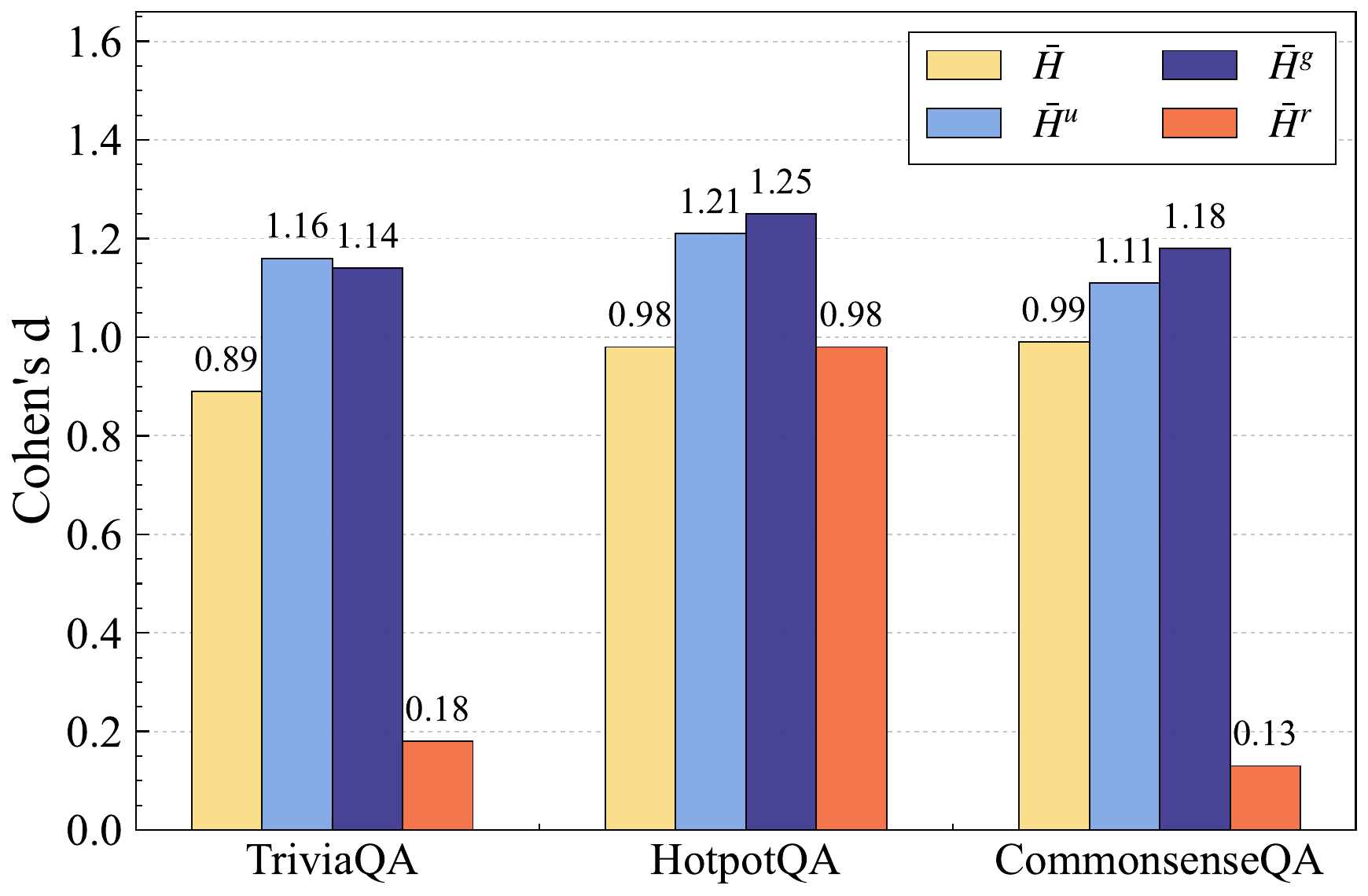}
    \caption{
    Evidence informativeness comparison on \llada with 128 diffusion steps. 
    We report Cohen's $d$ for all-token entropy $\bar H$, unrevealed-token entropy $\bar H^u$, revealing-token entropy $\bar H^g$, and revealed-token entropy $\bar H^r$ over the last 30\% of denoising steps. 
    Revealing-token entropy achieves the strongest separation on HotpotQA and CommonsenseQA, and remains highly competitive on TriviaQA.
    }
    \label{fig:app_evidence_llada_128}
\end{figure}

\paragraph{Analysis.}
Figure~\ref{fig:app_evidence_llada_128} provides complementary evidence to the main-text comparison under a longer denoising budget.
\ding{182}~Revealed-token entropy $\bar H^r$ is generally weak, especially on TriviaQA and CommonsenseQA, indicating that post-commitment residual uncertainty is not a reliable evidence source.
\ding{183}~All-token entropy $\bar H$ is consistently less discriminative than the best token-family evidence, suggesting that aggregating all positions dilutes the informative revelation-state structure.
\ding{184}~Unrevealed-token entropy $\bar H^u$ remains competitive, which is expected because it captures the difficulty of the survivor set.
However, revealing-token entropy $\bar H^g$ achieves the best Cohen's $d$ on two out of three datasets and is nearly tied with $\bar H^u$ on TriviaQA.
This supports our design choice that revealing tokens provide a commitment-aligned evidence source: they measure uncertainty at the moment when content enters the generated answer, rather than uncertainty that remains internal to the unresolved denoising process.
\subsection{Revealing-Token Entropy Trajectory}
\label{app:revealing_trajectory}

We provide additional revealing-token entropy trajectories across model families, datasets, and diffusion-step budgets.
For each denoising step $t$, we compute the revealing-token entropy mass
\begin{equation}
\widetilde H_t^g
=
\sum_{i\in T_t^g} H_t(i),
\label{eq:app_revealing_mass}
\end{equation}
where $T_t^g$ denotes the revealing-token set at step $t$.
For hallucinated and non-hallucinated samples, we plot the group-level trajectories
\begin{equation}
\mu_t^{(1)}
=
\mathbb E[\widetilde H_t^g \mid y=1],
\qquad
\mu_t^{(0)}
=
\mathbb E[\widetilde H_t^g \mid y=0],
\label{eq:app_group_traj}
\end{equation}
where $y=1$ and $y=0$ denote hallucinated and non-hallucinated responses, respectively.

\begin{figure*}[t]
    \centering
    \begin{subfigure}[t]{0.24\textwidth}
        \centering
        \includegraphics[width=\linewidth]{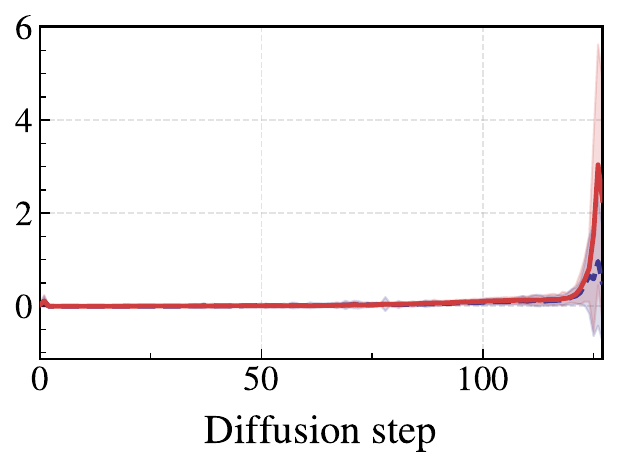}
        \caption{\llada on TriviaQA with 128 diffusion steps.}
        \label{fig:app_traj_llada_tqa128}
    \end{subfigure}
    \begin{subfigure}[t]{0.24\textwidth}
        \centering
        \includegraphics[width=\linewidth]{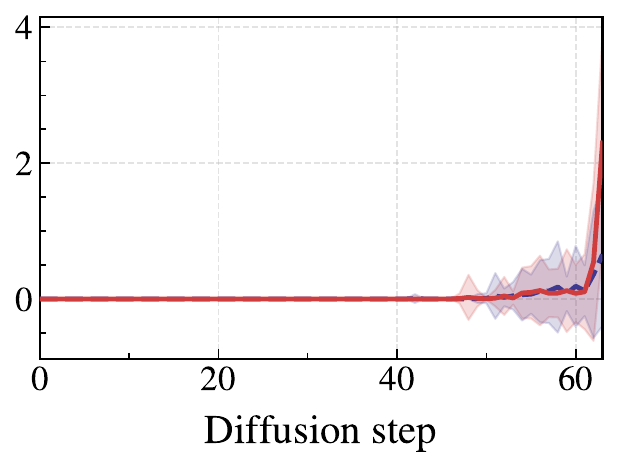}
        \caption{\dream on TriviaQA with 64 diffusion steps.}
        \label{fig:app_traj_dream_tqa64}
    \end{subfigure}
    \begin{subfigure}[t]{0.24\textwidth}
        \centering
        \includegraphics[width=\linewidth]{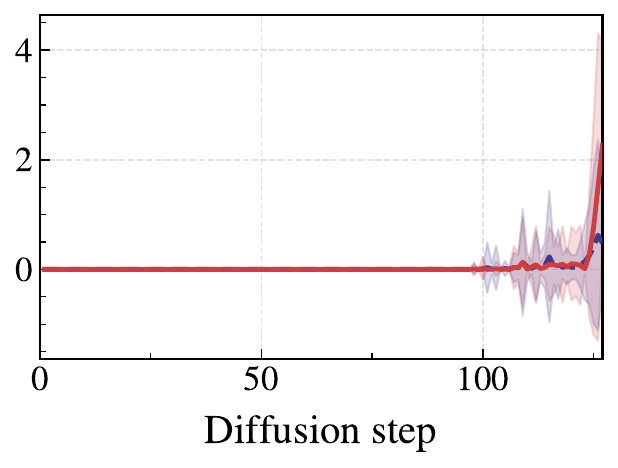}
        \caption{\dream on TriviaQA with 128 diffusion steps.}
        \label{fig:app_traj_dream_tqa128}
    \end{subfigure}
    \begin{subfigure}[t]{0.24\textwidth}
        \centering
        \includegraphics[width=\linewidth]{figures/method_evidence/method_traj/hotpotqa_steps64_newly_revealed_token_entropy_mean_std_dream.pdf}
        \caption{\dream on HotpotQA with 128 diffusion steps.}
        \label{fig:app_traj_dream_hqa128}
    \end{subfigure}

    \caption{
    Additional revealing-token entropy trajectories across model families, datasets, and diffusion-step budgets.
    Each plot shows the step-wise revealing-token entropy mass $\widetilde H_t^g$ for hallucinated and non-hallucinated samples.
    The separation between the two groups is generally concentrated near late denoising steps.
    }
    \label{fig:app_revealing_traj_extra}
\end{figure*}

\paragraph{Analysis.}
Figure~\ref{fig:app_revealing_traj_extra} provides additional evidence for the temporal behavior of revealing-token entropy mass.
\ding{182}~At early denoising steps, the revealing-token entropy mass remains small or weakly separated between hallucinated and non-hallucinated samples.
\ding{183}~Near the end of denoising, the hallucination/non-hallucination gap becomes more visible, showing that the discriminative signal is concentrated in late steps.
\ding{184}~Although the exact curve shape differs across model families and datasets, the late-stage separation is consistently observed across these additional settings.
These results support the temporal aggregation design of TRE, which assigns larger weights to later revealing-token entropy mass.
\section{Detailed Analogy of CDS and D-LLM}
\label{app:cds}
\input{appendix/cds}

\section{Detailed Experimental Setup}
\label{app:exp_setup}

\subsection{Datasets and Benchmarks}
\label{app:datasets}

We evaluate \ourmethod{} on three widely used question-answering benchmarks that require different forms of factual reasoning. These datasets allow us to examine whether a hallucination detector can remain effective across diverse answer-generation scenarios, ranging from direct factual recall to multi-hop reasoning and commonsense inference.

\begin{itemize}[leftmargin=*]
    \item \textbf{TriviaQA}~\cite{joshi2017triviaqa} is an open-domain question-answering dataset centered on factual knowledge. Since many questions require identifying specific entities, dates, locations, or events, this benchmark is useful for evaluating whether a detector can recognize failures in knowledge-intensive generations.
    
    \item \textbf{HotpotQA}~\cite{yang2018hotpotqa} contains questions that often require combining multiple pieces of evidence before producing the final answer. Compared with single-hop factual recall, this setting introduces additional reasoning complexity, making it suitable for testing whether hallucination detection methods can capture errors that arise during compositional or multi-step reasoning.
    
    \item \textbf{CommonsenseQA}~\cite{talmor2019commonsenseqa} focuses on commonsense reasoning. The questions typically require models to rely on implicit everyday knowledge rather than simply matching surface-level facts. This benchmark therefore complements the other two datasets by testing hallucination detection in scenarios where correctness depends on plausible world knowledge and commonsense associations.
\end{itemize}

\subsection{Model Backbones}
\label{app:model_backbones}

Our experiments are conducted on two representative diffusion-based large language models, \llada-8B-Instruct~\cite{nie2025large} and \dream-7B-Instruct~\cite{ye2025dream}. Both models generate responses through an iterative denoising process rather than the standard left-to-right autoregressive decoding mechanism. This makes them suitable testbeds for studying hallucination detection methods that exploit intermediate denoising behavior.

Using two different D-LLM backbones allows us to evaluate whether the proposed detector captures general properties of diffusion-based generation instead of overfitting to the trajectory characteristics of a single model~\cite{pan2026correcting}. For a given benchmark and backbone, all detectors are evaluated on the same set of generated responses whenever applicable, ensuring that performance differences mainly reflect the detectors themselves rather than variations in sampled answers.

\subsection{Baseline Configurations}
\label{app:baselines}

We compare \ourmethod{} with both training-based and training-free hallucination detection baselines. These methods differ in whether they require labeled examples for detector training and in which signals they use to estimate factual reliability.

\paragraph{Training-based Detectors.}
Training-based methods learn a mapping from model-derived features to correctness labels~\cite{tan2024influence,li2025treexformer}. In our experiments, they are trained separately for each benchmark using the corresponding automatically annotated training split, and are then evaluated on held-out examples from the same benchmark.

\begin{itemize}[leftmargin=*]
    \item \textbf{CCS}~\cite{burns2022discovering} learns a contrastive direction in the representation space by encouraging consistency between paired views of truthful and untruthful behavior. It provides a representative latent-probing baseline for detecting whether internal activations encode factual correctness.
    
    \item \textbf{TSV}~\cite{park2025steer} constructs a truthfulness-oriented separator vector from latent representations. The learned direction is then used to assign truthfulness scores to model outputs according to their positions in the hidden-state space.
    
    \item \textbf{TraceDet}~\cite{chang2025tracedet} is a trajectory-based detector designed for diffusion language models. Instead of relying only on the final answer, it analyzes information from the denoising process and aggregates trajectory-level signals for hallucination detection. Since TraceDet is a strong recent detector for D-LLMs, we treat it as the primary training-based baseline.
\end{itemize}

\paragraph{Training-Free Detectors.}
Training-free detectors estimate uncertainty or inconsistency directly from model outputs, token-level statistics, multiple sampled generations, or hidden representations~\cite{miao2026blindguard}.

\begin{itemize}[leftmargin=*]
    \item \textbf{Perplexity}~\cite{ren2022out} uses the likelihood assigned by the model to its generated sequence as a confidence signal. Responses with lower likelihood are generally treated as less reliable.
    
    \item \textbf{LN-Entropy}~\cite{malinin2020uncertainty} measures predictive uncertainty through length-normalized entropy. By averaging uncertainty over the generated sequence, it reduces the bias introduced by different response lengths.
    
    \item \textbf{Semantic Entropy}~\cite{kuhn2023semantic} estimates uncertainty from the semantic diversity of multiple sampled responses. Generations that express inconsistent meanings are assigned higher uncertainty, even if their surface forms differ only partially.
    
    \item \textbf{Lexical Similarity}~\cite{lin2023generating} evaluates the agreement among multiple sampled outputs using surface-level textual overlap. Lower similarity across samples suggests weaker generation stability and potentially higher hallucination risk.
    
    \item \textbf{EigenScore}~\cite{chen2024inside} derives a confidence score from the spectral properties of hidden representations. It captures variation in the internal representation space and uses this structure as an indicator of response reliability.
\end{itemize}

These methods are applied directly without additional optimization on the target datasets. For \dream-7B-Instruct, we do not report Perplexity and LN-Entropy because stable token-level logits are not reliably available in our experimental interface. This restriction prevents us from computing likelihood- or entropy-based scores in a consistent manner for that backbone.

\subsection{Implementation Details}
\label{app:implementation}

\paragraph{Dataset Construction.}
For each benchmark, we randomly select 800 question--answer pairs from the official evaluation split, following the scale used in recent hallucination detection studies. Each example contains a question, the corresponding reference answer, and the response generated by the evaluated D-LLM. When multiple reference answers are available, we keep all of them and regard a generated response as correct if it is semantically consistent with any valid reference answer.

\paragraph{Response Generation.}
For each model--dataset pair, we use a fixed prompt template and the same decoding configuration to generate responses. The generated responses are saved before applying any detector, ensuring that all compared methods are evaluated on identical model outputs rather than different sampled answers.

\paragraph{Decoding Configuration.}
We use deterministic decoding for response generation. Specifically, the temperature is set to $0$, while \texttt{top\_p} and \texttt{top\_k} are left unspecified, so no nucleus or top-$k$ sampling is applied. Models are loaded in \texttt{float16} precision, and \texttt{cfg\_scale} is set to $0.0$, meaning that no additional classifier-free guidance scaling is used. These settings are fixed across all datasets and backbones.

\paragraph{Automatic Annotation.}
To obtain correctness labels, we use \textbf{GPT-4o mini} as an automatic judge. Given a question, its reference answer, and the response generated by the evaluated D-LLM, GPT-4o mini is prompted to determine whether the generated response is semantically consistent with the reference answer. The judge is instructed to focus on factual correctness rather than surface-form overlap, so that paraphrases, abbreviations, or equivalent expressions are not incorrectly marked as wrong.

To improve label reliability, we adopt a conservative re-evaluation procedure. Responses initially judged as incorrect are sent to GPT-4o mini for a second judgment with the same question, reference answer, and generated response. If the two judgments are inconsistent, the example is discarded from the final evaluation set. After this filtering step, the remaining annotated examples are used as the final evaluation set for \ourmethod{} and all applicable baselines.

\paragraph{Feature Extraction and Scoring.}
\ourmethod{} does not require detector training. It directly computes hallucination detection scores from intermediate denoising information collected during the original D-LLM generation process. Thus, no task-specific parameter update or additional supervised training is performed for \ourmethod{}.

\paragraph{Computational Environment.}
All experiments are implemented with PyTorch and the HuggingFace Transformers library. Model inference, feature extraction, and detector scoring are conducted on a single NVIDIA Quadro RTX 6000 GPU with 24GB VRAM. Due to this single-GPU setting, experiments are run sequentially rather than in large-scale parallel batches.

We do not compare with~\cite{qian2026dynhd,hemmat2026tdgnet}, because their source code is not publicly available at the time of our experiments. Including these methods without official implementations would require nontrivial re-implementation choices and could introduce unfair or unreliable comparisons.

\section{Additional Experimental Results}

\subsection{Additional Ablation Study}
\label{app:additional_ablation}

We further examine the design choices on Dream to evaluate whether the observations generalize across D-LLM backbones. As shown in Table~\ref{tab:ablation_dream}, the results show similar trends to those on \llada. \ding{182}~Replacing newly revealed-token evidence with all-token, unrevealed-token, or revealed-token entropy consistently weakens the detector, confirming the importance of focusing on the reveal boundary. \ding{183}~For temporal aggregation, linear weighting performs competitively with exponential weighting and hard last-$30\%$ selection, while avoiding extra hyperparameters. \ding{184}~These results suggest that the effectiveness of \ourmethod does not depend on a specific backbone, and that late-weighted newly revealed-token entropy remains a robust signal for D-LLM hallucination detection.

\begin{table}[t]
\centering
\caption{Ablation study on Dream.}
\label{tab:ablation_dream}
\resizebox{0.45\textwidth}{!}{
\begin{tabular}{l|ccc}
\toprule
Variant & TriQA & HotQA & CSQA \\
\midrule
All tokens $\bar H$ & 74.6 & 73.8 & 63.9 \\
Unrevealed $\bar H^u$ & 76.4 & 75.7 & 60.2 \\
Revealed $\bar H^r$ & 62.1 & 62.9 & 63.4 \\
\midrule
Average & 83.2 & 79.7 & 77.4 \\
Exponential & \textbf{83.9} & 80.4 & \textbf{77.7} \\
Hard last-$30\%$ & 83.5 & 80.2 & \textbf{77.7} \\
\midrule
\ourmethod (ours) & \textbf{83.9} & \textbf{80.6} & \textbf{77.7} \\
\bottomrule
\end{tabular}
}
\end{table}

\begin{figure}[h]
\vspace{-0.8em}
\centering
\includegraphics[width=0.6\linewidth]{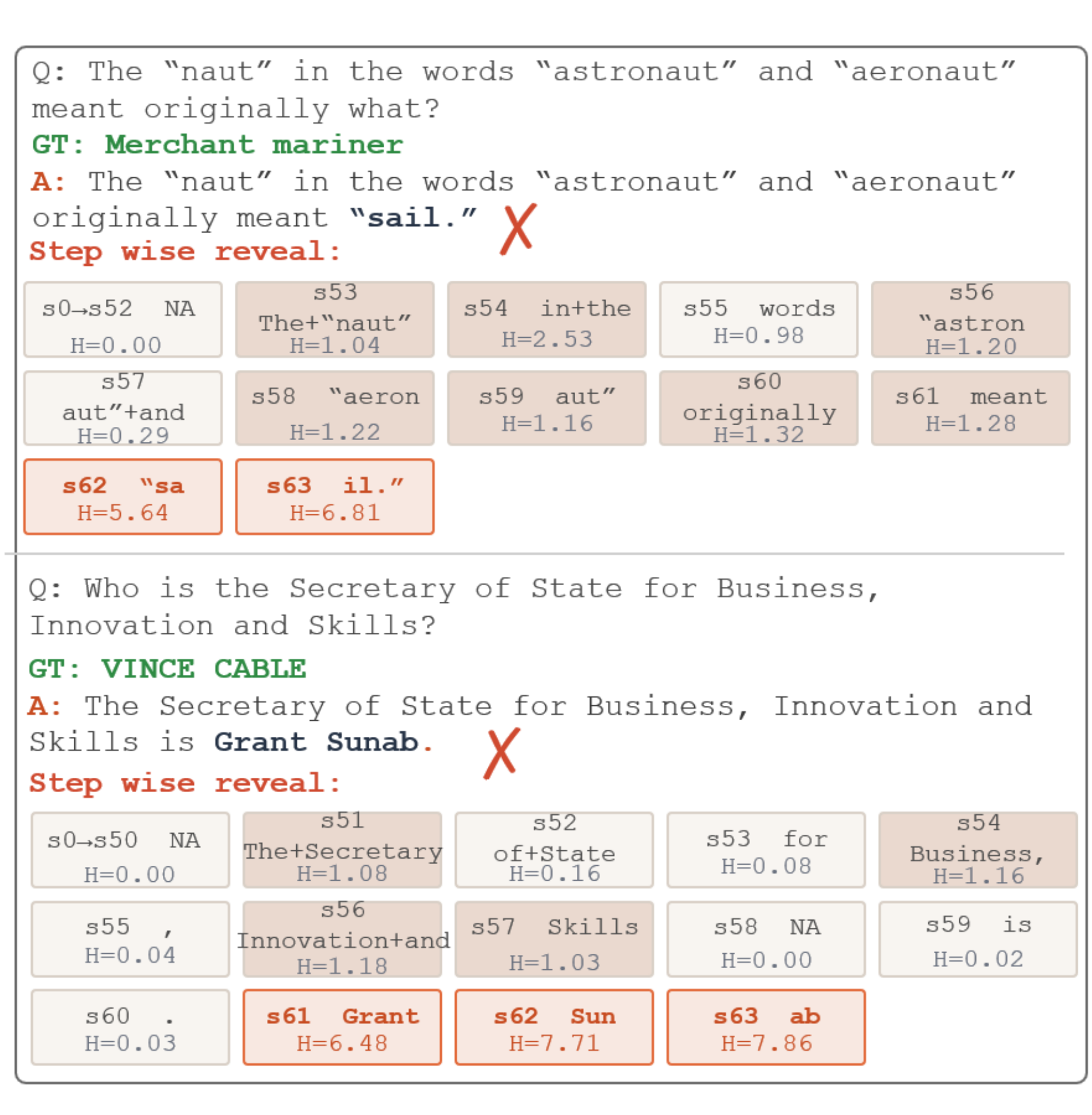}
\caption{Additional hallucinated case study examples. Hallucinated answer tokens are often newly revealed at late denoising steps and exhibit sharply increased entropy.}
\label{fig:appendix_case_study}
\vspace{-1.0em}
\end{figure}

\subsection{Additional Case Studies}
\label{app:case_study}
We provide additional hallucinated examples in Fig.~\ref{fig:appendix_case_study} to further illustrate the uncertainty dynamics captured by \ourmethod. Similar to the main case study, the generated tokens at each step (s) are listed with their revealing entropy (H) values. The results lead to the following observations. \ding{182}~Hallucinated answer tokens are consistently revealed at late denoising steps. For example, the model generates ``sail.'' instead of ``Merchant mariner'', and ``Grant Sunab'' instead of ``Vince Cable'', with these erroneous tokens appearing near the end of the denoising process. \ding{183}~When these hallucinated answers become newly revealed tokens, their entropy increases sharply, while earlier tokens usually have much smaller entropy. This further verifies that late-stage newly revealed-token entropy provides an informative and consistent signal for hallucination detection.

%% file: appendix/cds.tex
This appendix gives a stylized justification for why reveal-boundary entropy mass is a useful signal for hallucination detection in diffusion language models. The purpose is not to claim that D-LLM decoding obeys a literal thermodynamic law. Rather, we use a discrete diffusion and effective free-energy view to formalize a process-level intuition: D-LLM decoding repeatedly denoises a discrete token canvas while freezing selected positions through reveal events. Under this view, confidence-based reveal can create a hard residual reservoir, while revealing-token entropy mass measures the flux of this residual uncertainty into the committed answer. Incompatible commitments can further act as unstable or frustrated boundary conditions for the remaining unresolved subsystem.

\subsection{Discrete Diffusion, Reveal Operators, and Effective Energy}
\label{app:discrete_diffusion_energy}

We first cast D-LLM decoding as a discrete reveal-denoising process. Let $\mathcal V$ be the token vocabulary. At denoising step $t$, let $C_t$ denote the committed positions before the step, and let $V_t$ denote the active unresolved positions before reveal. In the notation of the main text,
\begin{equation}
C_t = R_{t-1},
\qquad
N_t = \Delta R_t,
\qquad
V_t = N_t \sqcup U_t ,
\label{eq:appendix_sets}
\end{equation}
where $N_t$ is the newly revealed subset at step $t$, and $U_t$ is the survivor set that remains unresolved after the reveal.

The model induces a predictive distribution over the active discrete token canvas,
\begin{equation}
q_t(x_{V_t}\mid x_{C_t}) \in \Delta(\mathcal V^{|V_t|}),
\label{eq:predictive_distribution_active}
\end{equation}
where $\Delta(\mathcal V^{|V_t|})$ denotes the probability simplex over token configurations on the active positions. This distribution represents the model's current belief over unresolved token configurations conditioned on the committed scaffold.

A D-LLM decoding step can be viewed as the composition of two operations. The first is a discrete denoising or reconditioning update, represented abstractly by an operator $K_t^\theta$. The second is a reveal operation, which selects a subset $N_t\subseteq V_t$ and freezes it to concrete token values $a_t$. At a process level,
\begin{equation}
q_{t+1}
\approx
K_t^\theta \circ Q^{\mathrm{reveal}}_{N_t,a_t}(q_t).
\label{eq:denoise_reveal_composition}
\end{equation}
Here $K_t^\theta$ reshapes the predictive distribution over the remaining unresolved positions, while $Q^{\mathrm{reveal}}_{N_t,a_t}$ removes degrees of freedom by conditioning on the revealed values.

For a fixed reveal set $N_t$ and fixed commitment $a_t$, the reveal operator acts as
\begin{equation}
\big(Q^{\mathrm{reveal}}_{N_t,a_t} q_t\big)(x_{U_t})
=
q_t(x_{U_t}\mid x_{C_t}, x_{N_t}=a_t),
\qquad
U_t = V_t\setminus N_t.
\label{eq:reveal_operator}
\end{equation}
Thus, reveal is a discrete projection/conditioning operation rather than merely a continuous noise-reduction step. In D-LLMs, uncertainty is not only denoised over time, but also crosses a moving reveal boundary where some variables become committed answer content.

The reveal subset is typically chosen by a local confidence rule. In an idealized form,
\begin{equation}
N_t
=
\operatorname{TopK}_{i\in V_t}
\big[-H_q(X_i\mid X_{C_t})\big],
\label{eq:topk_confidence_rule}
\end{equation}
and the committed token values are chosen by
\begin{equation}
a_{t,i}
=
\arg\max_{v\in\mathcal V}
q_t(X_i=v\mid X_{C_t}),
\qquad
i\in N_t.
\label{eq:local_commitment_rule}
\end{equation}
This rule is local: it selects positions whose marginal distributions appear confident under the current scaffold. It does not guarantee that the resulting commitment $a_t$ is globally optimal for the remaining configuration.

This discrete denoising-plus-reveal view motivates the rest of the analysis. The denoising operator $K_t^\theta$ updates the predictive distribution, while the reveal operator $Q^{\mathrm{reveal}}_{N_t,a_t}$ freezes selected degrees of freedom and changes the boundary condition for the remaining subsystem. Hallucination risk is therefore naturally tied to uncertainty at the reveal boundary rather than to a global average over all token positions.

Given the discrete predictive distribution above, we can introduce an effective energy representation. For any finite discrete distribution and any $\beta_t>0$, one may write
\begin{equation}
q_t(x_{V_t}\mid x_{C_t})
=
\frac{1}{Z_t(x_{C_t})}
\exp\{-\beta_t E_t(x_{V_t};x_{C_t})\},
\label{eq:gibbs_representation}
\end{equation}
where
\begin{equation}
E_t(x_{V_t};x_{C_t})
=
-\beta_t^{-1}\log q_t(x_{V_t}\mid x_{C_t})+\mathrm{const}.
\label{eq:effective_energy}
\end{equation}
This should be understood as an effective energy representation of the model's predictive distribution, not as a claim that D-LLM decoding is a physical thermodynamic system.

The corresponding residual entropy is
\begin{equation}
S_t
=
H_q(X_{V_t}\mid X_{C_t})
=
-\mathbb E_{q_t}\log q_t(X_{V_t}\mid X_{C_t}).
\label{eq:residual_entropy}
\end{equation}
The associated free energy is
\begin{equation}
F_t
=
-\beta_t^{-1}\log Z_t
=
\mathbb E_{q_t}[E_t]
-
\beta_t^{-1}S_t.
\label{eq:free_energy}
\end{equation}
The reveal operation is not a closed entropy-conserving evolution. It is an externally driven discrete quench: the protocol removes active degrees of freedom and changes the boundary condition seen by the remaining unresolved positions.

\subsection{Selection-Induced Residual Concentration and Boundary Entropy Flux}
\label{app:selection_boundary_flux}

Let
\begin{equation}
h_i(t)
=
H_q(X_i\mid X_{C_t})
\end{equation}
be the marginal uncertainty of position $i\in V_t$ before the reveal operation. Confidence-based reveal tends to select locally easier positions. We abstract this by assuming that the newly revealed subset has no larger average entropy than the active unresolved set:
\begin{equation}
\bar h(N_t)
:=
\frac{1}{|N_t|}
\sum_{i\in N_t}h_i(t)
\le
\frac{1}{|V_t|}
\sum_{i\in V_t}h_i(t)
=:
\bar h(V_t).
\label{eq:confidence_selection_assumption}
\end{equation}
Since $V_t=N_t\sqcup U_t$, this implies
\begin{equation}
\bar h(U_t)
:=
\frac{1}{|U_t|}
\sum_{i\in U_t}h_i(t)
\ge
\bar h(V_t).
\label{eq:survivor_concentration}
\end{equation}
Indeed, letting $\mu_N=\bar h(N_t)$, $\mu_U=\bar h(U_t)$, and $\mu_V=\bar h(V_t)$, we have
\begin{equation}
|V_t|\mu_V
=
|N_t|\mu_N
+
|U_t|\mu_U.
\end{equation}
Since $\mu_N\le\mu_V$,
\begin{equation}
|U_t|\mu_U
=
|V_t|\mu_V-|N_t|\mu_N
\ge
|V_t|\mu_V-|N_t|\mu_V
=
|U_t|\mu_V,
\end{equation}
and hence $\mu_U\ge\mu_V$.

This survivor-concentration effect explains why unresolved-token entropy can remain competitive: lower-entropy positions are preferentially frozen, leaving a survivor set enriched with harder positions. However, survivor concentration alone does not identify the best hallucination signal. The set $U_t$ still consists of unresolved variables; its uncertainty remains internal to the denoising process. Hallucination detection concerns uncertainty that is actually committed into the generated answer.

\paragraph{Reveal-Boundary Flux Identity.}
Define the active uncertainty mass, survivor uncertainty mass, and reveal-boundary entropy flux as
\begin{equation}
\mathcal E_t^{V}
:=
\sum_{i\in V_t}h_i(t),
\qquad
\mathcal E_t^{U}
:=
\sum_{i\in U_t}h_i(t),
\qquad
\Phi_t
:=
\sum_{i\in N_t}h_i(t).
\end{equation}
Because $V_t=N_t\sqcup U_t$, we have the exact identity
\begin{equation}
\mathcal E_t^{V}
=
\mathcal E_t^{U}
+
\Phi_t,
\qquad
\Phi_t
=
\mathcal E_t^{V}
-
\mathcal E_t^{U}.
\label{eq:flux_identity}
\end{equation}
Thus, $\Phi_t$ is exactly the uncertainty mass that crosses the reveal boundary at step $t$.

\paragraph{Why This Flux is Diagnostic.}
Let $\ell_i$ denote the unobserved factual error loss associated with committing position $i$ into the answer, and let
\begin{equation}
r_i(t)
:=
\mathbb E[\ell_i\mid X_{C_t}, i\in N_t]
\end{equation}
be its conditional commitment risk at step $t$. Since the detector is training-free and entropy-only, $r_i(t)$ is not directly observable. A standard uncertainty-risk assumption is that commitment risk is monotone in predictive uncertainty:
\begin{equation}
r_i(t)
\approx
\psi(h_i(t)),
\qquad
\psi'(\cdot)\ge 0.
\label{eq:entropy_risk_proxy}
\end{equation}
Under the simplest local linear approximation $\psi(h)\propto h$, the total commitment risk at step $t$ satisfies
\begin{equation}
\sum_{i\in N_t}r_i(t)
\propto
\sum_{i\in N_t}h_i(t)
=
\Phi_t.
\label{eq:flux_risk_proxy}
\end{equation}
Therefore, $\Phi_t$ is the natural entropy-only proxy for the amount of risky content committed at the reveal boundary.

The mass form is important. The average $\bar h(N_t)$ measures uncertainty density per newly committed token, whereas
\begin{equation}
\Phi_t
=
|N_t|\bar h(N_t)
\label{eq:mass_density_relation}
\end{equation}
also accounts for the volume of content committed at the step. A trajectory can be risky either because a few committed tokens are highly uncertain or because many moderately uncertain tokens are committed together. The entropy mass captures both effects.

The resulting interpretation is
\begin{equation}
\text{confidence-based reveal}
\Longrightarrow
\text{hard residual reservoir}
\Longrightarrow
\text{diagnostic entropy flux when revealed}.
\label{eq:selection_to_flux_chain}
\end{equation}
Unresolved entropy explains where difficulty is stored; revealing-token entropy mass measures how much of that difficulty is committed into the answer.

\subsection{Pointwise Entropy Amplification under Discrete Reveal}
\label{app:pointwise_entropy_amplification}

The previous subsection shows that confidence-based selection can create a hard residual reservoir, while reveal-boundary entropy flux records when residual uncertainty becomes committed answer content. We now explain why a specific reveal event can amplify downstream uncertainty, even though conditioning reduces entropy on average.

For a candidate reveal value $a$ on $N_t$, define the post-commitment residual entropy
\begin{equation}
S_t(a)
=
H_q(X_{U_t}\mid X_{C_t}, X_{N_t}=a).
\label{eq:post_commitment_entropy}
\end{equation}
A standard conditional entropy identity gives
\begin{equation}
\mathbb E_{a\sim q(X_{N_t}\mid X_{C_t})}
S_t(a)
=
H_q(X_{U_t}\mid X_{C_t}, X_{N_t}),
\end{equation}
and hence
\begin{equation}
\mathbb E_a S_t(a)
=
H_q(X_{U_t}\mid X_{C_t})
-
I_q(X_{N_t};X_{U_t}\mid X_{C_t}).
\label{eq:conditioning_entropy_identity}
\end{equation}
Since mutual information is nonnegative,
\begin{equation}
\mathbb E_a S_t(a)
\le
H_q(X_{U_t}\mid X_{C_t}).
\label{eq:average_conditioning_reduces_entropy}
\end{equation}
Thus, conditioning on reveal values reduces residual entropy on average.

However, decoding does not average over all possible values of $a$. It commits to a specific value $a_t$, often chosen by a local confidence rule. Define the pointwise entropy jump
\begin{equation}
\Delta S_t(a_t)
=
H_q(X_{U_t}\mid X_{C_t},X_{N_t}=a_t)
-
H_q(X_{U_t}\mid X_{C_t}).
\label{eq:pointwise_entropy_jump}
\end{equation}
Also define the pointwise excess term
\begin{equation}
\epsilon_t(a_t)
=
H_q(X_{U_t}\mid X_{C_t},X_{N_t}=a_t)
-
\mathbb E_a H_q(X_{U_t}\mid X_{C_t},X_{N_t}=a).
\label{eq:pointwise_excess}
\end{equation}
Then
\begin{equation}
\Delta S_t(a_t)
=
\epsilon_t(a_t)
-
I_q(X_{N_t};X_{U_t}\mid X_{C_t}).
\label{eq:jump_excess_mi}
\end{equation}
Therefore,
\begin{equation}
\Delta S_t(a_t)>0
\quad\Longleftrightarrow\quad
\epsilon_t(a_t)>I_q(X_{N_t};X_{U_t}\mid X_{C_t}).
\label{eq:positive_jump_condition}
\end{equation}
This shows that average entropy reduction under conditioning is compatible with entropy amplification after a specific greedy commitment. A token may be locally confident under the current scaffold while still being globally incompatible with the remaining configuration, producing a positive pointwise residual entropy jump. In such cases, reveal-boundary entropy mass is a useful observable: it records uncertain commitments at the step where they become answer content.

\subsection{Committed Tokens as Attention-Induced Boundary Conditions}
\label{app:attention_boundary_conditions}

Committed tokens do not disappear after being revealed. In a diffusion language model with bidirectional attention, committed positions continue to influence unresolved positions. We therefore treat committed tokens as boundary variables for the active unresolved subsystem.

Let $A^{(\ell,h,t)}_{ij}$ be the attention weight from position $i$ to position $j$ in layer $\ell$, head $h$, and step $t$. Define the averaged attention kernel
\begin{equation}
\kappa^{(t)}_{ij}
=
\frac{1}{LH}
\sum_{\ell=1}^{L}
\sum_{h=1}^{H}
A^{(\ell,h,t)}_{ij}.
\label{eq:attention_kernel}
\end{equation}
Since attention is not necessarily symmetric, we use the symmetrized kernel
\begin{equation}
\widetilde\kappa^{(t)}_{ij}
=
\frac{1}{2}
\left(
\kappa^{(t)}_{ij}
+
\kappa^{(t)}_{ji}
\right)
\label{eq:sym_attention_kernel}
\end{equation}
when writing an energy-like interaction functional.

Let $z_i$ be a continuous relaxation of the token state at position $i$, for example an embedding or a probability vector. We use the following stylized interaction functional as an abstraction of bidirectional contextual coupling:
\begin{equation}
E_t(z_{V_t};z_{C_t})
=
\sum_{i\in V_t} V_i(z_i)
+
\sum_{i,j\in V_t}
\widetilde\kappa^{(t)}_{ij}\Psi(z_i,z_j)
+
\sum_{\substack{i\in V_t\\ j\in C_t}}
\widetilde\kappa^{(t)}_{ij}\Psi(z_i,z_j).
\label{eq:attention_energy}
\end{equation}
The final term is the coupling between unresolved positions and committed positions. It represents the boundary field induced by the committed scaffold.

A coherent scaffold provides compatible boundary conditions, which can sharpen the conditional distributions of the remaining positions. A hallucinated or globally incompatible scaffold may instead impose conflicting constraints. To formalize this intuition, define a frustration functional after commitment:
\begin{equation}
J_t(C_t)
=
\min_{z_{U_t}}
\left[
\sum_{i,j\in U_t}
\widetilde\kappa^{(t)}_{ij}
d^2(z_i,T_{ij}z_j)
+
\sum_{\substack{i\in U_t\\j\in C_t}}
\widetilde\kappa^{(t)}_{ij}
d^2(z_i,T_{ij}z_j)
\right],
\label{eq:frustration_functional}
\end{equation}
where $T_{ij}$ is a compatibility transform and $d(\cdot,\cdot)$ is a distance in the relaxed state space. Small $J_t(C_t)$ means the unresolved variables can satisfy the imposed constraints coherently. Large $J_t(C_t)$ indicates a frustrated boundary condition.

\subsection{Boundary Susceptibility and Unstable Scaffolds}
\label{app:boundary_susceptibility}

We now connect boundary frustration with stability. Consider a smooth relaxed free-energy landscape
\begin{equation}
F_t(z_{U_t};z_{C_t})
=
E_t(z_{U_t};z_{C_t})
-
\tau_t S(z_{U_t}),
\label{eq:relaxed_free_energy}
\end{equation}
where $E_t$ is a relaxed energy, $S$ is an entropy functional over unresolved states, and $\tau_t>0$ is a temperature-like parameter.

Assume that, for fixed committed variables $z_{C_t}$, the unresolved subsystem is near a local stationary solution:
\begin{equation}
\nabla_{z_{U_t}}F_t(z^\star_{U_t};z_{C_t})=0.
\label{eq:stationary_condition}
\end{equation}
Let
\begin{equation}
\mathcal H_t
=
\nabla^2_{z_{U_t}z_{U_t}}
F_t(z^\star_{U_t};z_{C_t}),
\qquad
\mathcal G_t
=
\nabla^2_{z_{U_t}z_{C_t}}
F_t(z^\star_{U_t};z_{C_t}).
\label{eq:hessian_cross}
\end{equation}

\paragraph{Proposition 1.}
Assume that $F_t$ is twice continuously differentiable in a neighborhood of $(z^\star_{U_t},z_{C_t})$, that Equation~\eqref{eq:stationary_condition} holds, and that $\mathcal H_t$ is nonsingular. Then, for a sufficiently small boundary perturbation $\delta z_{C_t}$, the induced change in the local stationary solution satisfies
\begin{equation}
\delta z^\star_{U_t}
=
-\mathcal H_t^{-1}\mathcal G_t\delta z_{C_t}
+
O(\|\delta z_{C_t}\|^2).
\label{eq:stationary_response}
\end{equation}
Consequently, the boundary susceptibility is controlled by
\begin{equation}
\chi_t
=
\|\mathcal H_t^{-1}\mathcal G_t\|.
\label{eq:boundary_susceptibility}
\end{equation}
If $\mathcal H_t\succeq \gamma I$ for some $\gamma>0$, then
\begin{equation}
\|\delta z^\star_{U_t}\|
\le
\frac{\|\mathcal G_t\|}{\gamma}
\|\delta z_{C_t}\|
+
O(\|\delta z_{C_t}\|^2).
\label{eq:stable_bound}
\end{equation}
When $\lambda_{\min}(\mathcal H_t)$ is close to zero, the susceptibility can become large. If $\mathcal H_t$ has a negative eigenvalue, the stationary point is unstable.

\paragraph{Proof.}
The stationary condition is
\begin{equation}
\nabla_{z_U}F_t(z^\star_U(z_C);z_C)=0.
\end{equation}
Perturb $z_C$ to $z_C+\delta z_C$ and the stationary solution to $z^\star_U+\delta z^\star_U$. A first-order Taylor expansion gives
\begin{align}
0
&=
\nabla_{z_U}F_t(z^\star_U+\delta z^\star_U;z_C+\delta z_C)\\
&=
\nabla_{z_U}F_t(z^\star_U;z_C)
+
\mathcal H_t\delta z^\star_U
+
\mathcal G_t\delta z_C
+
O(\|\delta z_C\|^2).
\end{align}
The first term is zero by stationarity. Therefore,
\begin{equation}
\mathcal H_t\delta z^\star_U+\mathcal G_t\delta z_C
=
O(\|\delta z_C\|^2).
\end{equation}
Since $\mathcal H_t$ is nonsingular,
\begin{equation}
\delta z^\star_U
=
-\mathcal H_t^{-1}\mathcal G_t\delta z_C
+
O(\|\delta z_C\|^2).
\end{equation}
Taking norms gives
\begin{equation}
\|\delta z^\star_U\|
\le
\|\mathcal H_t^{-1}\mathcal G_t\|
\|\delta z_C\|
+
O(\|\delta z_C\|^2).
\end{equation}
If $\mathcal H_t\succeq\gamma I$, then $\|\mathcal H_t^{-1}\|\le 1/\gamma$, and hence Equation~\eqref{eq:stable_bound} follows. If $\mathcal H_t$ has a negative eigenvalue, the stationary point is not a local minimum of the free energy and is therefore unstable.
\hfill$\square$

This proposition gives a precise meaning to stable and unstable committed scaffolds. A coherent commitment corresponds to a well-conditioned basin, where small boundary changes have limited effect on the unresolved subsystem. An incompatible commitment can create a flat, metastable, or unstable scaffold, in which later denoising updates are amplified through the inverse Hessian.

If the entropy functional is locally smooth, then
\begin{equation}
\delta S
=
\langle \nabla_{z_U}S,\delta z^\star_U\rangle
+
O(\|\delta z^\star_U\|^2).
\label{eq:entropy_response}
\end{equation}
Thus, an unstable scaffold can induce a large entropy response whenever the boundary perturbation has a component in an entropy-increasing direction. This formalizes the intuition that a hallucinated commitment may appear locally confident at the moment of reveal but later destabilize the residual subsystem.

\subsection{From Boundary Entropy Flux to TRE}
\label{app:boundary_flux_to_tre}

The analysis above suggests that the most diagnostic uncertainty is not the bulk entropy stored in the unresolved reservoir, but the entropy mass that crosses the reveal boundary and becomes committed answer content. Let $N_t=\Delta R_t$ denote the newly revealed positions at step $t$. We define the reveal-boundary entropy flux as
\begin{equation}
\Phi_t
=
\sum_{i\in N_t}H_t(i).
\label{eq:boundary_flux_final}
\end{equation}
A large $\Phi_t$ means that the decoder is committing either many moderately uncertain tokens or a smaller number of highly uncertain tokens. In both cases, $\Phi_t$ measures the total amount of uncertain content entering the answer at step $t$.

TRE aggregates this reveal-boundary flux with a monotone late-stage weight:
\begin{equation}
\mathrm{TRE}(x)
=
\sum_{t=1}^{T}
w\!\left(\frac{t}{T}\right)
\Phi_t,
\qquad
w(u)=u.
\label{eq:tre_flux}
\end{equation}
Equivalently,
\begin{equation}
\mathrm{TRE}(x)
=
\sum_{t=1}^{T}
\frac{t}{T}
\sum_{i\in N_t}H_t(i).
\label{eq:tre_flux_expanded}
\end{equation}
This is the revealing-token entropy mass score used in the main text. It measures the total amount of uncertain commitment crossing the reveal boundary, with stronger emphasis on late denoising.

The mass form is intentional. An average over newly revealed tokens measures uncertainty density per committed token, while the mass $\Phi_t$ measures the total amount of uncertain content committed at that step. Hallucination risk can depend on both: a trajectory is risky not only when individual commitments are uncertain, but also when a large volume of uncertain content is committed late. This should not be interpreted as a count-only effect. The reveal count $|N_t|$ measures how much content is committed, while the entropy values $H_t(i)$ measure how uncertain those commitments are. Their product-like mass aggregation captures total uncertain commitment crossing the reveal boundary.

Thus, TRE measures persistent reveal-boundary entropy flux: uncertainty that is committed into the answer and weighted more strongly when it occurs late in denoising.

\subsection{Summary of the Formal Picture}
\label{app:formal_summary}

The stylized analysis supports the detector design through the following chain:
\begin{align}
&\text{discrete denoising + reveal projection}
&&\Longrightarrow
\text{a moving commitment boundary},
\\
&\text{confidence-based reveal}
&&\Longrightarrow
\text{a hard residual reservoir},
\\
&\text{reveal events}
&&\Longrightarrow
\text{residual uncertainty crosses into the answer},
\\
&\text{committed tokens}
&&\Longrightarrow
\text{boundary conditions through bidirectional attention},
\\
&\text{wrong commitments}
&&\Longrightarrow
\text{frustrated or unstable residual subsystem},
\\
&\text{unstable residual subsystem}
&&\Longrightarrow
\text{late entropy amplification at the reveal boundary}.
\end{align}

This formal picture does not claim a universal physical law of D-LLM hallucination. Rather, it explains the detector structure: unresolved entropy describes where residual difficulty is stored, while revealing-token entropy mass measures how much of that difficulty is committed into the answer. The proposed score therefore tracks reveal-boundary entropy flux and emphasizes late commitments, where uncertain content is less likely to be revised away.

%% file: sections/Limitations_and_Broader_Impacts.tex
\section{Limitations and Broader Impacts}
\label{sec:lim}

While TRE provides a simple and efficient training-free metric for hallucination detection in diffusion language models, it still has several limitations. First, TRE is based on entropy signals from the denoising process and does not explicitly use semantic knowledge. Therefore, it may be less effective when a model produces incorrect answers with high confidence. Second, our experiments are conducted on question-answering benchmarks with two representative D-LLM backbones. Evaluating TRE on more tasks, domains, and future diffusion language models would further verify its generality.

This work may help improve the reliability of diffusion language models by providing a lightweight hallucination detection signal without additional detector training or repeated generation. However, TRE should be used as an auxiliary indicator rather than a guarantee of factual correctness. In practical applications, especially in high-stakes scenarios, its predictions should be combined with human review or external verification when necessary. This paper releases only an anonymized supplementary code package for the detector and does not release new datasets, model checkpoints, or high-risk generative models.